\documentclass{article}

\usepackage[preprint]{neurips_2026}
\usepackage{amsmath}
\usepackage{multirow}
\usepackage{amssymb}

\usepackage{graphicx} 
\usepackage{mathabx}
\usepackage{subcaption}
\usepackage{xcolor}

\usepackage{arydshln}

\usepackage[utf8]{inputenc} 
\usepackage[T1]{fontenc}    
\usepackage{hyperref}       
\usepackage{url}            
\usepackage{booktabs}       
\usepackage{amsfonts}       
\usepackage{nicefrac}       
\usepackage{microtype}      
\usepackage{xcolor}         

\title{When Does Discrete Ricci Curvature Help? An Edge-Level Analysis of Curvature-Enhanced GCNs for PPI Prediction}
\workshoptitle{Geometric Distributional Deep Learning}

\author{%
Clara Joswig\\
Department of Mathematics, ETH Zurich\\
Zurich, Switzerland\\
\texttt{cjoswig@student.ethz.ch}\\
\AND
Elham Gholamzadeh\\
Max Planck Institute for Human Cognitive and Brain Sciences\\ 
Leipzig, Germany\\
\texttt{gholamzade@cbs.mpg.de}\\
\AND
Ghazal Mostofi\\
Max Planck Institute for Human Cognitive and Brain Sciences\\ 
Leipzig, Germany\\
\texttt{mostofi@cbs.mpg.de}\\
\AND
Nico Scherf\\
Max Planck Institute for Human Cognitive and Brain Sciences\\
Center for Scalable Data Analytics and Artificial Intelligence (ScaDS.AI)\\
Leipzig, Germany
\texttt{nscherf@cbs.mpg.de}\\
}

\begin{document}

\maketitle

\begin{abstract}
Curvature-enhanced graph convolutional networks (CGCN) incorporate Ollivier--Ricci curvature (ORC), grounded in optimal transport theory, into message passing, motivated by the claim that curvature-weighted edges alleviate "over-squashing" at structural bottlenecks. But does CGCN's improvement actually come from mitigating over-squashing? We investigate this on four real-world protein-protein interaction (PPI) datasets, comparing CGCN against a standard GCN at the level of individual edges rather than aggregate metrics alone. While CGCN improves aggregate AUC, AUPR, and F1 across all four datasets, this gain is concentrated almost entirely on low-degree edges, while performance on low-curvature (bottleneck) edges systematically degrades in three of four datasets, the opposite of what the over-squashing motivation predicts. Examining the learned edge-weight function directly, together with clustering-validity metrics, edge-distance ratios, and a layer-depth ablation across four depths, we find evidence for an alternative account: curvature-weighting sharpens the separation between clusters, which improves intra-cluster predictions at the cost of inter-cluster (bottleneck) ones. We test this account against topologically defined communities rather than the classification labels used elsewhere.
Additionally, we offer practical guidance on when to prefer CGCN over standard GCN, depending on whether intra- or inter-cluster predictions matter most.
\end{abstract}

\section{Introduction}
Interactions between biomolecules--proteins, drugs, genes, and microbes--often involve multiple binding partners acting cooperatively, and understanding these interactions remains a central problem in computational biology \cite{moll2006biomolecular}. Many such systems, along with larger-scale networks like social, citation, and transportation networks, can naturally be represented as graphs, making Graph Neural Networks (GNNs) a natural tool for modeling them. Standard architectures such as GCN \cite{kipf2016semi}, GAT \cite{velickovic2017graph}, and GIN \cite{xu2019powerful} rely on neighborhood aggregation over directly connected nodes.


Geometric deep learning embeds structural priors of this kind directly into network
architectures rather than treating them as post-hoc descriptors \cite{bronstein2021geometric,
atz2021geometric}. One such prior is discrete Ricci curvature, most commonly formalized for
graphs via Ollivier--Ricci curvature (ORC) \cite{ollivier2009ricci} (grounded
in optimal transport theory) or Forman--Ricci curvature (FRC) \cite{sreejith2016forman}. ORC characterizes each edge according to its neighborhood's
connectivity: edges in densely connected, highly clustered regions tend to have positive
curvature, reflecting redundant alternative paths, while edges spanning sparsely connected
regions have negative curvature, reflecting a lack of such alternatives — these are
structural bottlenecks \cite{topping2021understanding}.

One way to leverage discrete Ricci curvature on graphs is via discrete Ricci flow \cite{ni2019community}, which iteratively adjusts edge weights or lengths based on local curvature differences to reshape the network's internal geometry. This technique finds applications in geometric embeddings \cite{ni2018network}, community detection \cite{ni2019community}, or graph rewiring methods \cite{topping2021understanding}.

Alternatively, discrete Ricci curvature can be incorporated into GNNs directly to mitigate structural failure modes. Over-squashing occurs when information from distant nodes must repeatedly traverse such bottleneck edges across multiple rounds of message passing, causing it to be increasingly compressed and distorted as it propagates \cite{topping2021understanding}. A related failure mode, over-smoothing, arises in densely connected regions, where repeated averaging across redundant paths causes node representations to become increasingly similar and less discriminative \cite{chen2020measuring}. 
Because negatively curved edges coincide with the regions where over-squashing is expected to concentrate \cite{topping2021understanding}, curvature has been proposed as a signal for identifying and correcting for these bottlenecks via graph rewiring (e.g., SDRF \cite{topping2021understanding}, FoSR \cite{karhadkar2023fosr}), curvature-based positional encodings, or curvature-weighted message passing \cite{yadav2024roadmap} (e.g., CGCN \cite{shen2024curvature}, CurvGN \cite{ye2020curvature}, CGNN \cite{li2022curvature}). Here, we focus on the curvature-weighted message passing paradigm, instantiated by CGCN \cite{shen2024curvature}, which incorporates ORC directly into the message-passing operator to weight edges by local geometry.

Curvature-based descriptors have already proven useful directly on PPI networks: early work applying Forman- and Ollivier--Ricci curvature to human and yeast PPI networks showed that curvature distributions reveal structural heterogeneity not captured by degree alone \cite{eidi2020edge}, and curvature-informed features have been combined with machine learning for protein-ligand interaction prediction \cite{wee2021forman}. Building on this, curvature-enhanced GNNs incorporate ORC directly into the message-passing operator, differentially weighting edges by local geometric context. CGCN \cite{shen2024curvature} is one such model, and has been shown to outperform standard GCNs across several real-world biomolecular interaction prediction tasks, including PPI.


In this work, we investigate the influence of ORC in message passing empirically on four real-world PPI datasets, comparing a standard GCN against CGCN not just in terms of aggregate performance, but at the level of individual edges, with the goal of understanding the mechanism behind CGCN and the role ORC plays within it. We claim that CGCN uses the additional ORC information in message passing to achieve better cluster separation within the embedding space. This leads to improved overall performance, but at the cost of degrading performance on low-curvature (i.e., bottleneck) edges. Our contribution is threefold: (1) we confirm that CGCN improves aggregate metrics (AUC, AUPR, F1) across all four datasets, consistent with \cite{shen2024curvature}; (2) we show this improvement is not uniform — it concentrates almost entirely on low-degree edges, while performance on low-curvature (bottleneck) edges systematically degrades under CGCN in three of four datasets, the opposite of what the over-squashing motivation would predict; and (3) by examining the learned edge-weight function directly and testing it against clustering-validity metrics, inter/intra-cluster distance ratios, and a layer-depth ablation, we provide empirical support for the following explanation: curvature-weighting sharpens cluster separation in the learned embeddings, improving prediction of intra-cluster edges at the cost of the inter-cluster (low-curvature) edges themselves.

\section{Graph Neural Networks}

We represent protein-protein interactions as a graph $G = (V, E)$, where nodes correspond to
proteins and edges to interactions. The neighborhood of $v \in V$ is $N(v) := \{u \mid \{u,v\}
\in E\}$, and each node $v_i$ has a feature vector $x_i$. We use a Graph Convolutional Network
(GCN), which updates node representations via message passing:
\begin{equation}
    x_i^l = \phi\left(x_i^{l-1}, \bigoplus_{j: v_j \in N(v_i)} c_{ij}\, \psi\left(x_j^{l-1}\right)\right),
    \label{eq:gnn-abstract}
\end{equation}
where $\bigoplus$ is an aggregation function and $c_{ij}$ weights the importance of $v_j$ to
$v_i$ \cite{bronstein2021geometric}. Following \cite{shen2024curvature}, we use the concrete
realization
\begin{equation}
    x_i^l = \sigma\left(\sum_{j: v_j \in N(v_i) \cup \{v_i\}}
    \frac{1}{\sqrt{\deg(v_i)}\sqrt{\deg(v_j)}} \cdot f(v_i, v_j) \cdot W \cdot x_j^{l-1}\right),
    \label{eq:gcn-concrete}
\end{equation}
with summation aggregation, degree normalization, learnable weight matrix $W$, and edge-weight function
$f(v_i, v_j)$ used to incorporate curvature (defined in Section~\ref{sec:models}).

\subsection{Ollivier--Ricci Curvature}
In Riemannian geometry, curvature measures how a manifold deviates from being locally Euclidean, with Ricci curvature specifically quantifying this deviation in tangent directions \cite{riemann2016hypotheses, samal2018comparative}.
Ollivier--Ricci curvature (ORC) extends Ricci curvature to graphs via optimal transport
\cite{ollivier2009ricci}: for each edge, it compares graph distance between endpoints to the
transport distance between probability distributions over their neighborhoods. For any node $v \in V$ with non-empty neighborhood $N(v) \neq \emptyset$, we define
\begin{align*}
    m_v(w) = \begin{cases}
        \alpha, & \text{if } w = v \\
        \dfrac{1-\alpha}{|N(v)|}, & \text{if } w \in N(v) \\
        0, & \text{otherwise},
    \end{cases}
\end{align*}
with $\alpha \in [0,1]$ (here we use $\alpha = 0.5$). For isolated nodes where $N(v) = \emptyset$, we set $m_v(v)=1$ and $m_v(w)=0$ for $w\neq v$.
Moreover, the 1-Wasserstein distance $W_1$ is defined as follows:
\begin{align*}
     W_1(m_u, m_v) = \inf_{\pi \in \Pi(m_u, m_v)} \sum_{x\in V}\sum_{y \in V} \pi(x,y)\, d(x,y),
\end{align*}
where $\Pi(m_u, m_v)$ denotes the set of all valid transport plans (couplings) with marginal distributions $m_u$ and $m_v$, and $d(x,y)$ represents the shortest-path distance between nodes $x$ and $y$.
This allows us to define the ORC of edge $\{u,v\}\in E$, denoted by $\kappa_{u,v}$, as follows:
\begin{align*}
    \kappa_{u,v} = 1 - \frac{W_1(m_u, m_v)}{d(u,v)}.
\end{align*}
Edges with
$\kappa_{u,v} < 0$ act as structural bottlenecks; $\kappa_{u,v} > 0$ indicates dense, clustered
regions \cite{topping2021understanding}.

\subsection{Models}
\label{sec:models}

We compare two variants differing only in $f(v_i, v_j)$ from Equation~(\ref{eq:gcn-concrete}). The
baseline GCN sets $f(v_i, v_j) = 1$. CGCN \cite{shen2024curvature} instead learns an edge weight
$f_c$ depending on ORC $\kappa_{u,v}$:
\begin{align*}
    G_c(u,v) &= \left(\frac{1+e^{-\kappa_{u,v}}}{2}, \dots, \frac{1+e^{-N \cdot \kappa_{u,v}}}{2}\right)^T, \\
    f_c(u,v) &= \tau\left(W_{\text{MLP}}^T G_c(u,v) + b\right),
\end{align*}

with $N = 10$ \cite{shen2024curvature}, activation $\tau(\cdot)$, and learnable
$W_{\text{MLP}} \in \mathbb{R}^{N \times 1}$, $b\in \mathbb{R}$. CGCN is obtained by setting
$f(u,v) = f_c(u,v)$ in Equation~(\ref{eq:gcn-concrete}).
For both models, edge predictions use
\begin{align*}
    \hat{p}_{v_i v_j} = \text{MLP}\big(\|(x_i + x_j,\; x_i \odot x_j,\; x_i,\; x_j)\big),
\end{align*}
with $\odot$ the element-wise product and $\|(\cdot)$ concatenation \cite{shen2024curvature}. See
\cite{shen2024curvature} for comparisons against other state-of-the-art methods.

\section{Materials}
\subsection{Datasets}\label{dataset}
We consider four protein-protein interaction datasets: HuRI-PPI, PPT-Ohmnet, PP-Pathways, and
PP-Decagon. All four are obtained from \cite{shen2024curvature}, who cite \cite{HuRI} as the
source for HuRI-PPI and PPT-Ohmnet (open access, CC BY 4.0), and \cite{PP}
as the source for PP-Pathways and PP-Decagon (publicly available for academic use).

HuRI-PPI was generated by systematically testing all
pairwise protein combinations in an all-by-all binary interaction space \cite{HuRI}. In contrast,
the other three datasets are built upon experimentally validated physical interactions from
primary databases, with PP-Pathways and PPT-Ohmnet further incorporating biological pathway
annotations and tissue-specific gene expression data, respectively \cite{PPT-Ohmnet,
PP-Pathways}. This reliance on curated primary databases can introduce dense clustering driven
by study bias and complex-level data representation \cite{HuRI, rolland2014}; HuRI-PPI, by
contrast, is systematically screened and therefore largely free of such bias \cite{HuRI}. To
verify this structural difference, we report clustering-structure metrics, namely: average clustering
coefficient, transitivity, modularity, and number of communities  in
Table~\ref{tab:cluster_structure} (Appendix~\ref{app:datasets}). Despite this potential study
bias, PPT-Ohmnet, PP-Pathways, and PP-Decagon contain valuable information, particularly given
their much larger size relative to HuRI-PPI. We emphasize these methodology-driven structural
differences to account for the contrasting behavior of HuRI-PPI relative to the other three
datasets in our subsequent evaluations. Dataset properties are summarized in
Table~\ref{tab:metrics} (Appendix~\ref{app:datasets}).

\subsection{Implementation Details}
\label{sec:implementation_details}
All experiments were conducted on an HPC system, equipped with Intel Xeon IceLake-SP
processors and NVIDIA A100 GPU nodes interconnected via NVLink. Our implementation builds
closely on that of \cite{shen2024curvature}.
Negative samples are selected randomly to match the number of positive samples, and the whole
dataset is divided into training, validation, and test sets in a 7:1:2 ratio. We run each test over 10 random seeds and report the mean and standard
deviation as final results. The run of 10 random seeds in the described environment took between one hour and several days, depending on the dataset. As in \cite{shen2024curvature}, we use no external node features,
relying instead on the adjacency matrix as the feature matrix. Moreover, as in \cite{shen2024curvature} we use a one-layer implementation of CGCN and GCN, respectively, and a batch size of 128 with
the Adam optimizer at a learning rate of 5e-4, and implement both CGCN and GCN in PyTorch. For
more detailed parameter settings, we refer to the original source code as presented in \cite{shen2024curvature}\footnote{Source code available
at: https://github.com/CS-BIO/CGCN.} (licensed under MIT).

The exploratory data analysis experiments reported in Section~\ref{sec:results} were conducted separately, using the same system, on two different kinds of DataFrames including the train and test set, respectively, enriched by additional information generated during the run of CGCN, GCN, respectively. The compute time of the experiments is negligible. However, more experiments were conducted, which we choose not to include due to missing relevance.

\section{Results}
\label{sec:results}
We first confirm that our reimplementation reproduces the aggregate
performance gains reported by \cite{shen2024curvature}: CGCN improves over GCN in loss, AUC,
AUPR, and F1 across all four datasets (full results in Appendix~\ref{app:baseline}). This
aggregate improvement, however, is exactly what motivates our central question: \emph{which}
edges drive this gain, and does the resulting pattern support the over-squashing motivation
typically given for curvature-weighting? We address this next.

Comparing which specific edges each model predicts correctly, when both are evaluated on
identical train, validation, and test splits, reveals that CGCN also performs \emph{worse} than
GCN on a subset of edges (see Table~\ref{tab:summary}).

\begin{table}[ht]
  \centering
  \begin{tabular}{lcccc}
    \hline
    Metric & HuRI-PPI & PPT-Ohmnet & PP-Pathways & PP-Decagon \\
    \hline
    GCN {\color{red}$\times$} CGCN {\color{red}$\times$} & $11.48\% \pm 0.44\%$ & $14.08\% \pm 0.47\%$ & $9.51\% \pm 0.28\%$ & $10.58\% \pm 0.25\%$ \\
    GCN {\color{green!60!black}$\checkmark$} CGCN {\color{green!60!black}$\checkmark$} & $83.08\% \pm 0.42\%$ & $79.07\% \pm 0.44\%$ & $84.33\% \pm 0.35\%$ & $80.81\% \pm 0.15\%$ \\
    GCN {\color{red}$\times$} CGCN {\color{green!60!black}$\checkmark$} & $3.16\% \pm 0.42\%$ & $4.19\% \pm 0.56\%$ & $3.95\% \pm 0.37\%$ & $6.00\% \pm 0.32\%$ \\
    GCN {\color{green!60!black}$\checkmark$} CGCN {\color{red}$\times$} & $2.29\% \pm 0.36\%$ & $2.67\% \pm 0.33\%$ & $2.22\% \pm 0.21\%$ & $2.61\% \pm 0.12\%$ \\
    \hline
  \end{tabular}
  \caption{Percentage of test-set edges (positive and negative samples combined) falling into
  each of the four correctness categories (\textcolor{red}{$\times$} = incorrect,
  \textcolor{green!60!black}{$\checkmark$} = correct), averaged over
  10 random seeds (mean $\pm$ standard deviation), comparing GCN and CGCN on identical
  train/validation/test splits.}
  \label{tab:summary}
\end{table}

The two models agree on most edges: both predict correctly on roughly 79--84\% of edges,
and both miss a further $\sim$9--14\%. The remaining edges, however, reveal a meaningful
disagreement: CGCN correctly predicts 3--6\% of edges that GCN misses, while GCN
correctly predicts roughly 2--3\% that CGCN misses. This asymmetry shows that the aggregate gains
in Table~\ref{tab:performance} (Appendix~\ref{app:baseline}) reflect a genuine shift in \emph{which} edges are predicted
correctly, not a uniform improvement across all edges.

\begin{figure}[ht]
    \centering
    \begin{subfigure}[b]{0.45\textwidth}
        \centering
        \includegraphics[width=\textwidth]{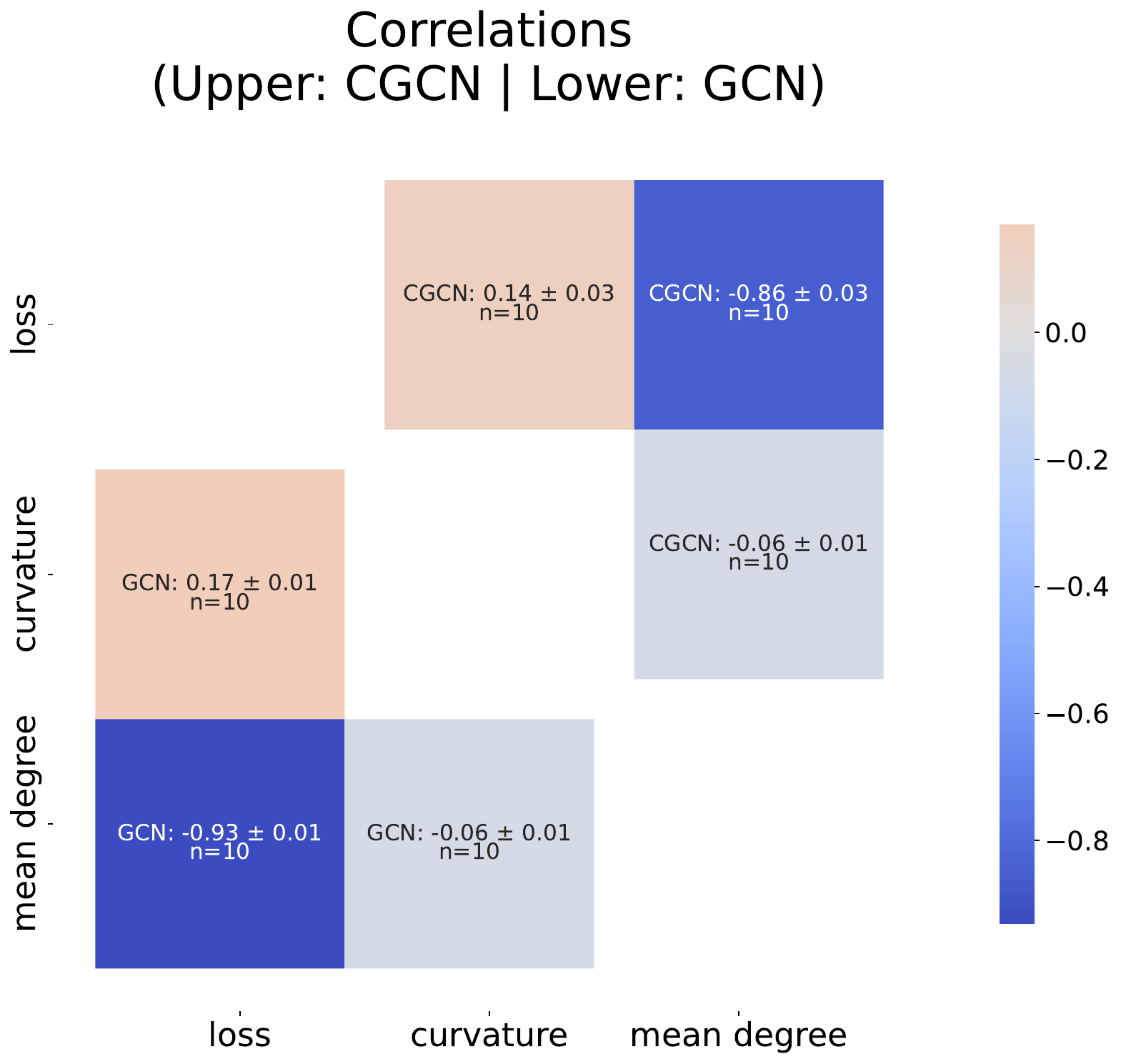}
        \caption{HuRI-PPI}
        \label{fig:bild1}
    \end{subfigure}
    \hfill
    \begin{subfigure}[b]{0.45\textwidth}
        \centering
        \includegraphics[width=\textwidth]{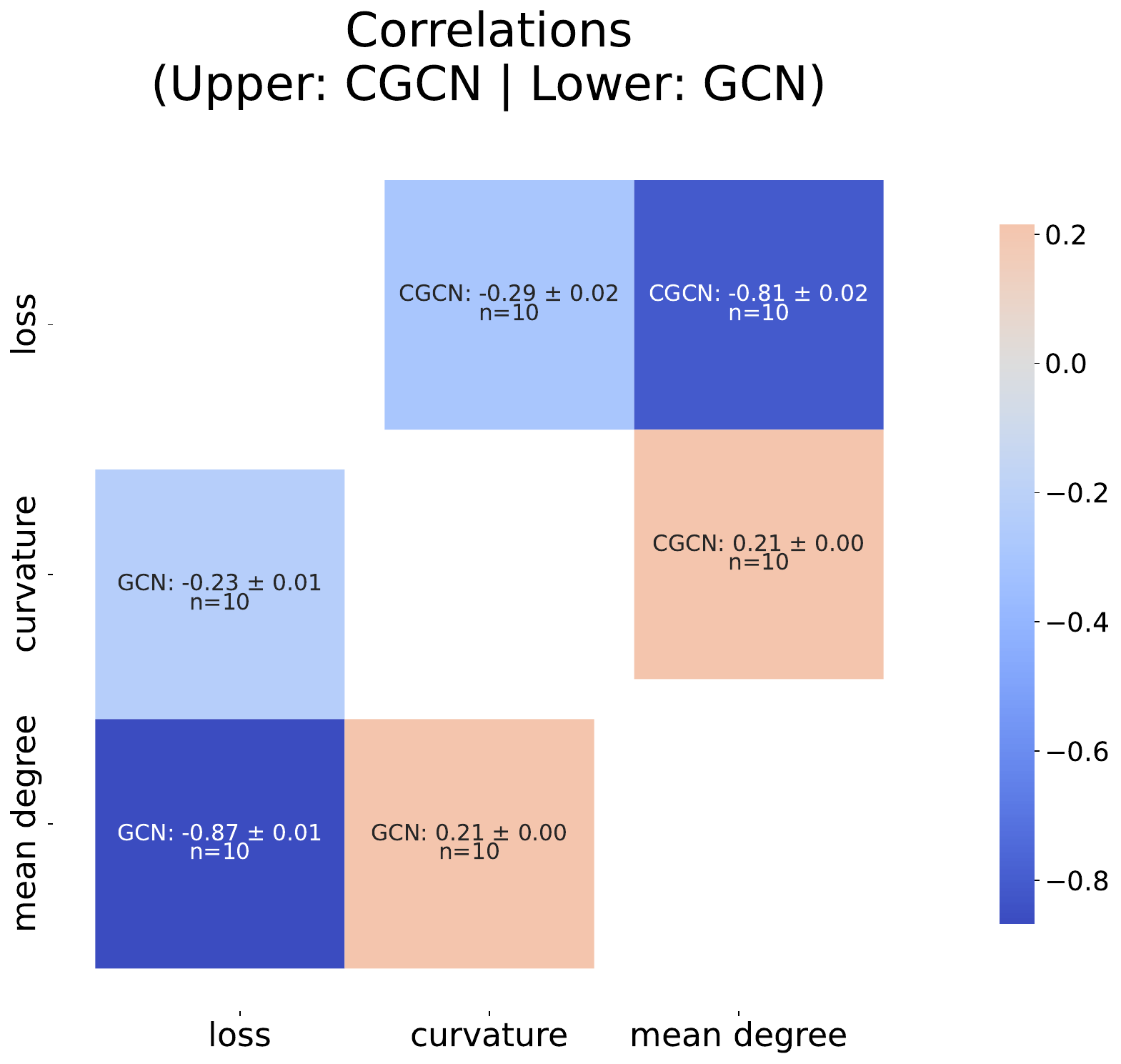}
        \caption{PPT-Ohmnet}
        \label{fig:bild2}
    \end{subfigure}

    \vspace{0.5cm}
    \begin{subfigure}[b]{0.45\textwidth}
        \centering
        \includegraphics[width=\textwidth]{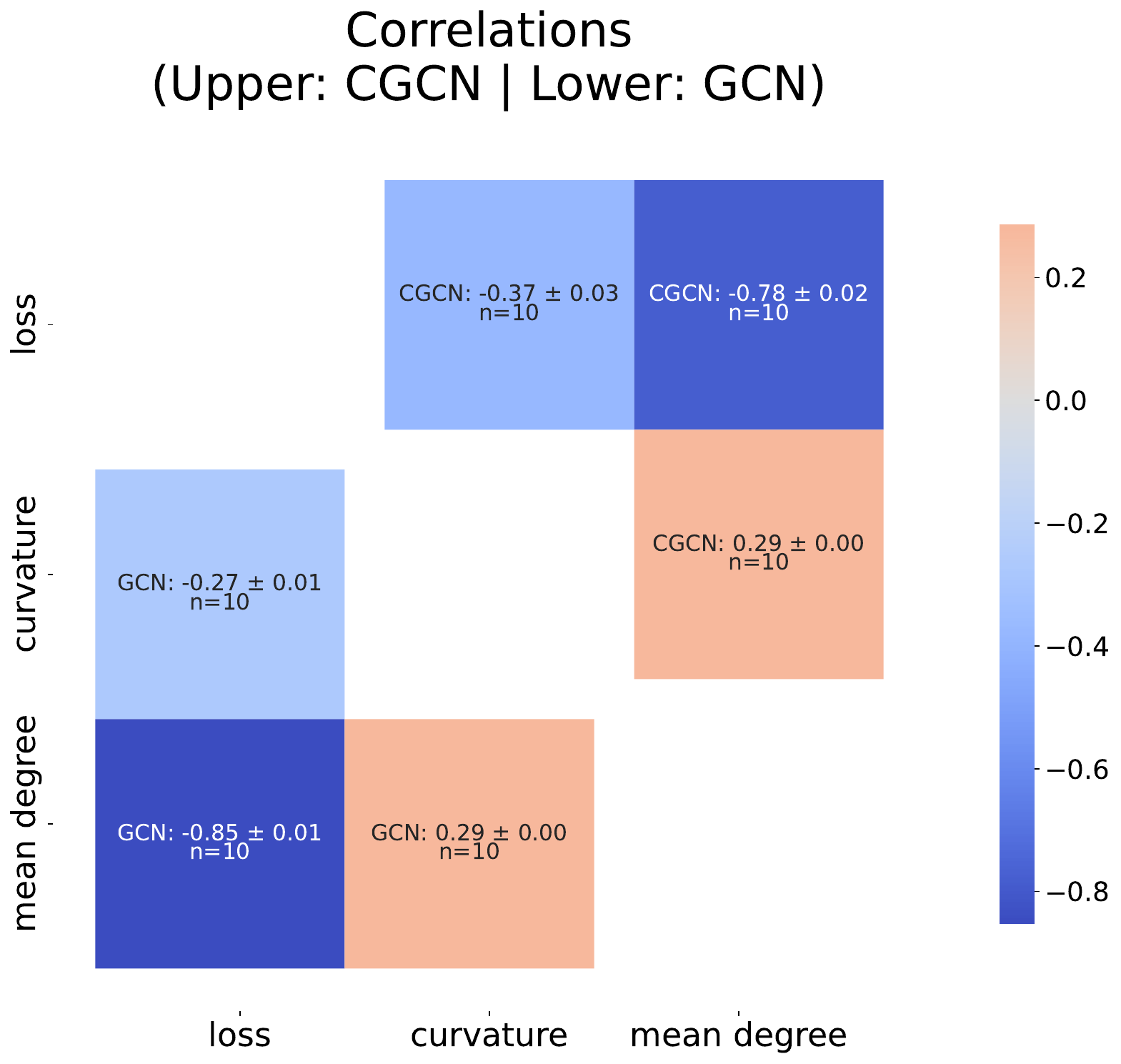}
        \caption{PP-Pathways}
        \label{fig:bild3}
    \end{subfigure}
    \hfill
    \begin{subfigure}[b]{0.45\textwidth}
        \centering
        \includegraphics[width=\textwidth]{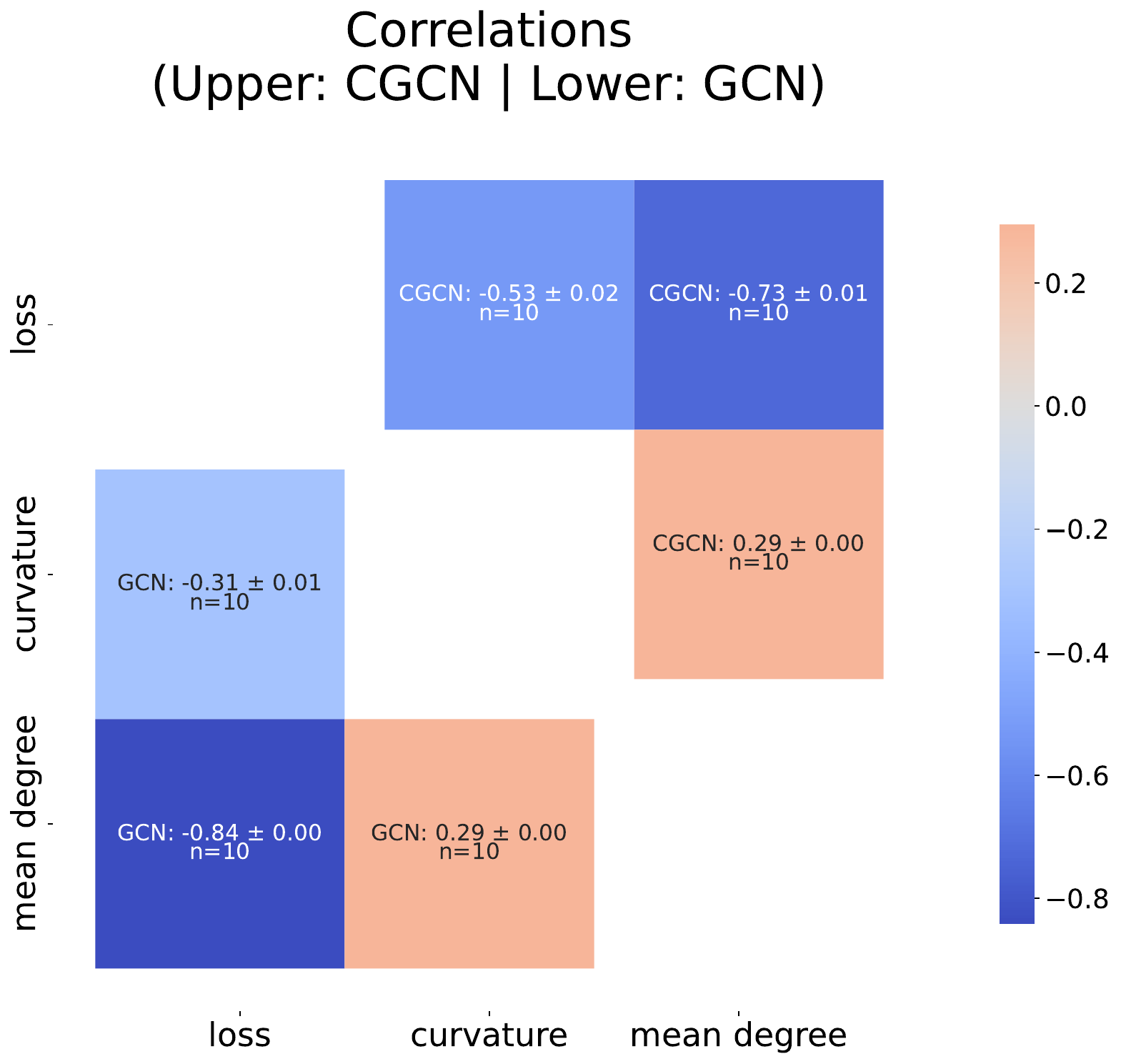}
        \caption{PP-Decagon}
        \label{fig:bild4}
    \end{subfigure}
    \caption{Correlations between edge loss, curvature, and mean node degree, for GCN and CGCN,
    across the four PPI datasets, averaged over
  10 random seeds (mean $\pm$ standard deviation).}
    \label{fig:corr}
\end{figure}

For the remainder of the analysis we restrict to positive edges, i.e., edges that actually
exist in the graph since these correspond to real biological interactions and are of
primary interest for downstream applications. (Negative edges were included above only to
report overall discriminative performance.) To identify what drives this shift, we
characterize each edge by its incident nodes' mean degree and its ORC value; both are negatively correlated with loss for both models, with HuRI-PPI being the only exception.
(Figure~\ref{fig:corr}).

\begin{figure}[ht]
    \centering
    \begin{subfigure}[b]{0.45\textwidth}
        \centering
        \includegraphics[width=\textwidth]{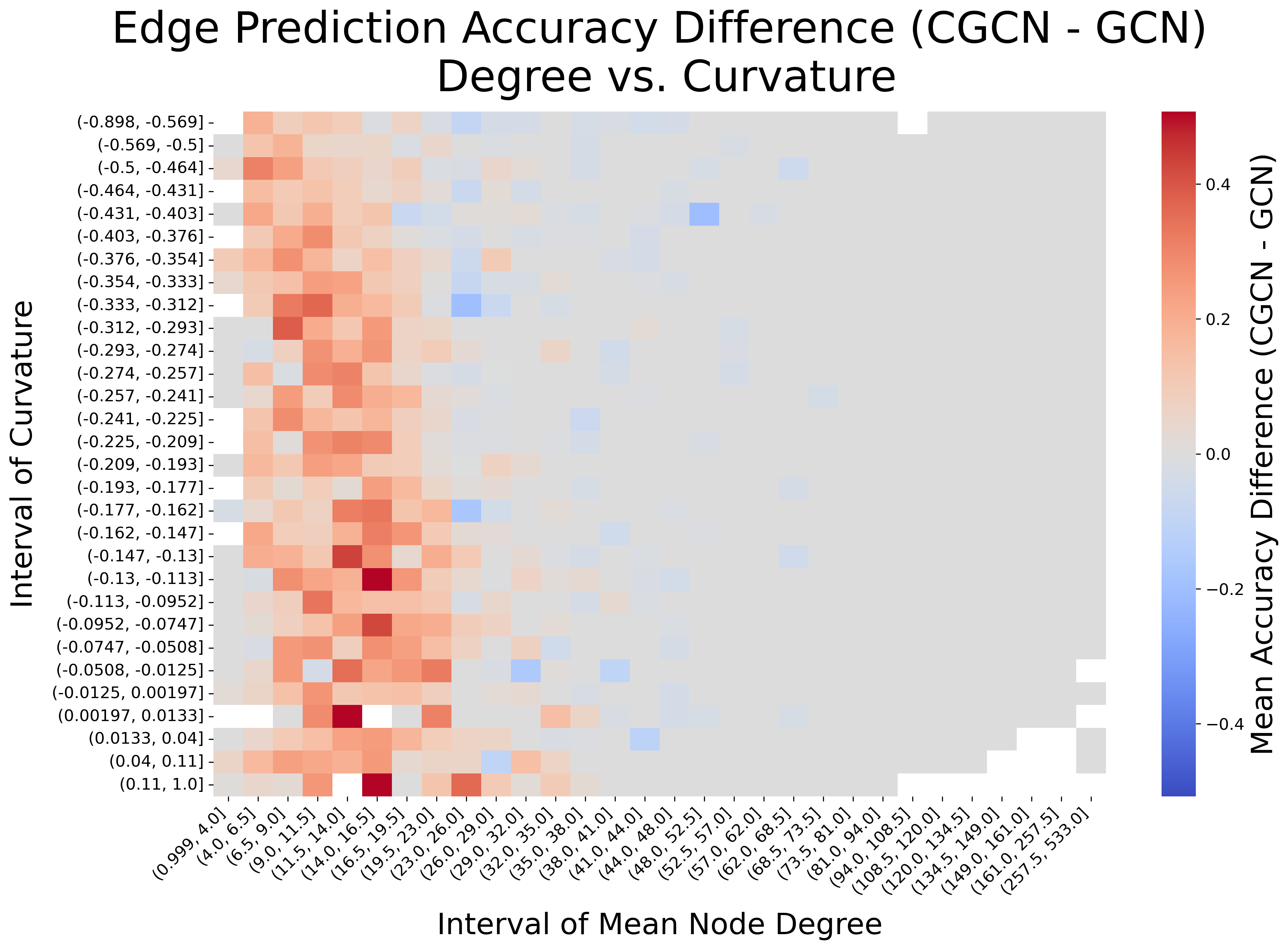}
        \caption{HuRI-PPI}
        \label{fig:heat-huri}
    \end{subfigure}
    \hfill
    \begin{subfigure}[b]{0.45\textwidth}
        \centering
        \includegraphics[width=\textwidth]{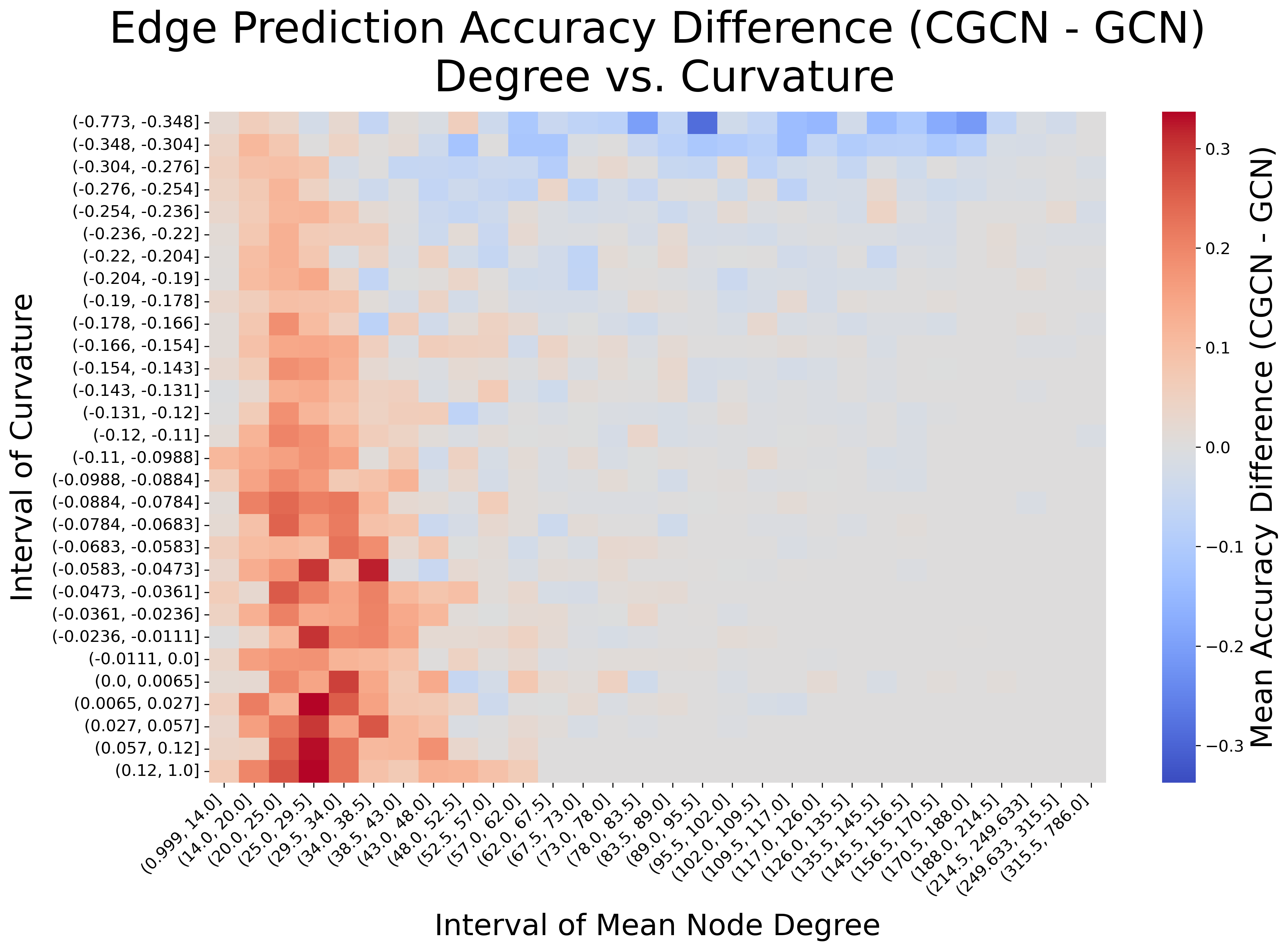}
        \caption{PPT-Ohmnet}
        \label{fig:heat-ohmnet}
    \end{subfigure}

    \vspace{0.5cm}
    \begin{subfigure}[b]{0.45\textwidth}
        \centering
        \includegraphics[width=\textwidth]{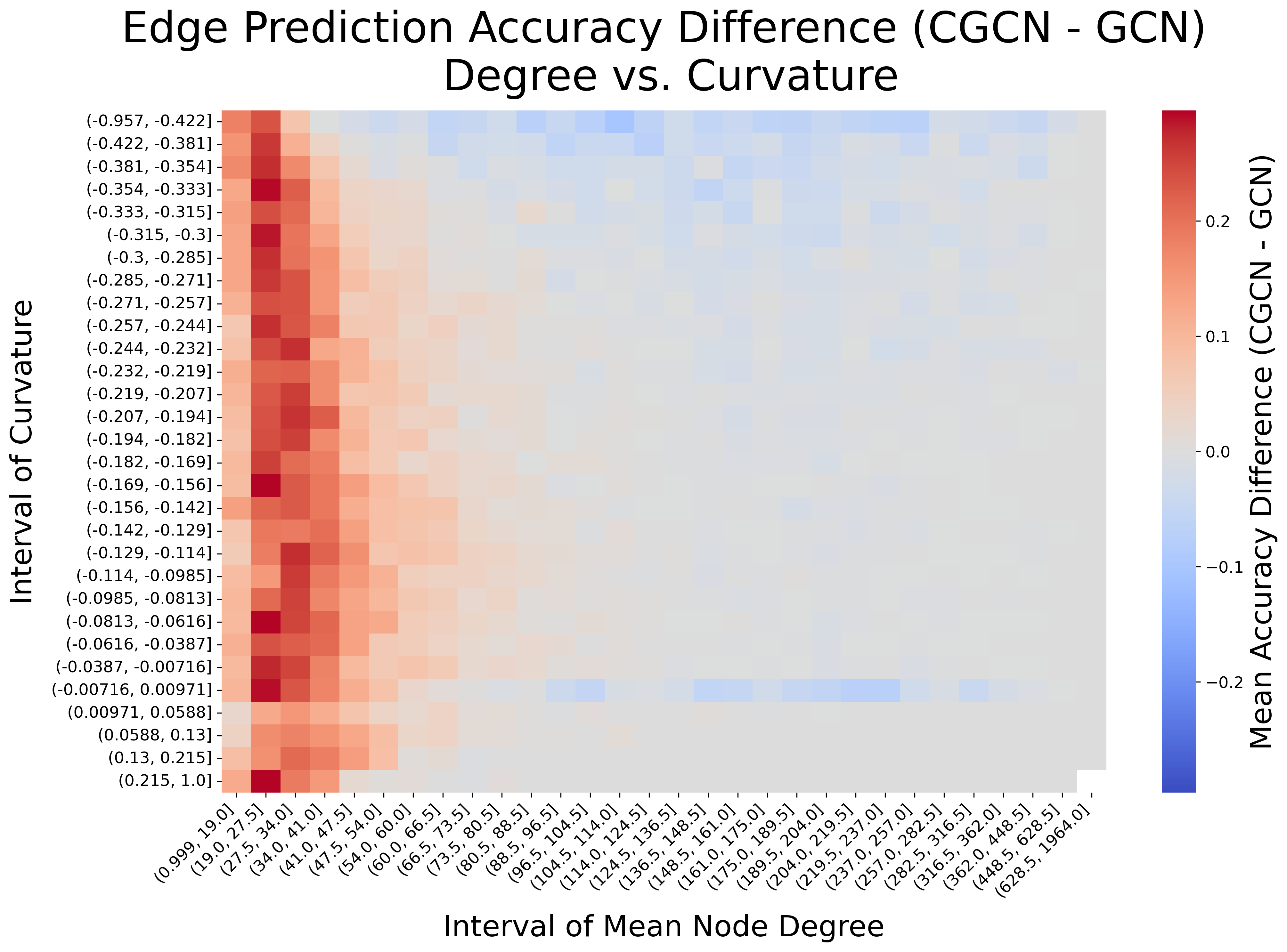}
        \caption{PP-Pathways}
        \label{fig:heat-pathways}
    \end{subfigure}
    \hfill
    \begin{subfigure}[b]{0.45\textwidth}
        \centering
        \includegraphics[width=\textwidth]{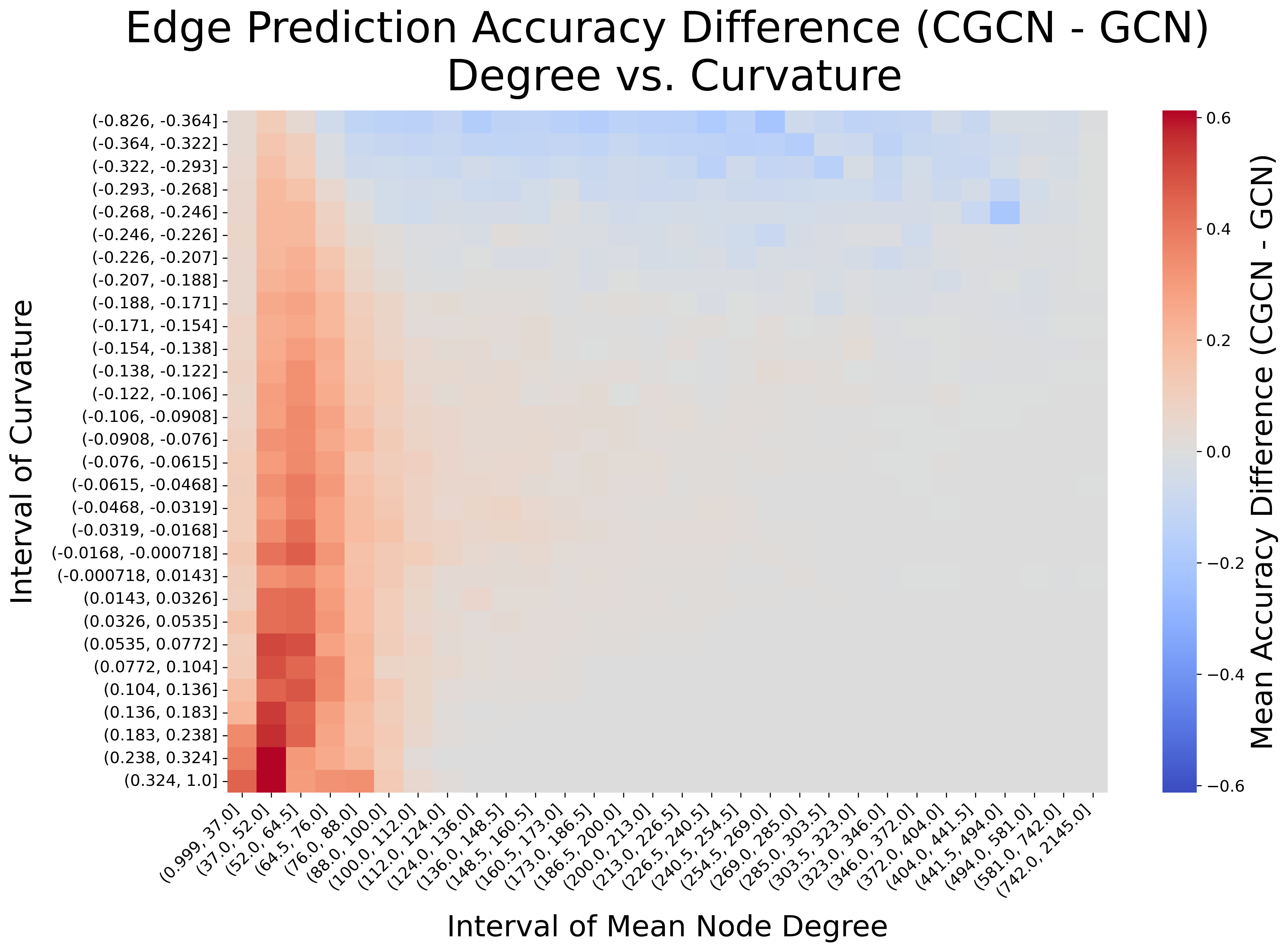}
        \caption{PP-Decagon}
        \label{fig:heat-decagon}
    \end{subfigure}
    \caption{Accuracy difference (CGCN $-$ GCN) by curvature and mean-degree bin, across the
    four PPI datasets, averaged over 10 random seeds.} 
    \label{fig:heat}
\end{figure}

The negative correlation between loss and mean degree is stronger for GCN, while the correlation between loss and ORC decreases, shifting toward lower (more negative) values, for CGCN, for all datasets, including HuRI-PPI. To visualize this, we bin edges into 30 equifrequent bins (equal edge count per bin), sorted separately by ORC and by mean degree, and report the accuracy difference between CGCN and GCN for each resulting bin combination (Figure~\ref{fig:heat}).

We find that CGCN improves over GCN for low-mean-degree edges, but yields a degradation
concentrated in the lowest-curvature edges in most datasets, with HuRI-PPI being a notable
exception.

\begin{figure}[ht]
    \centering
    \begin{subfigure}[b]{0.45\textwidth}
        \centering
        \includegraphics[width=\textwidth]{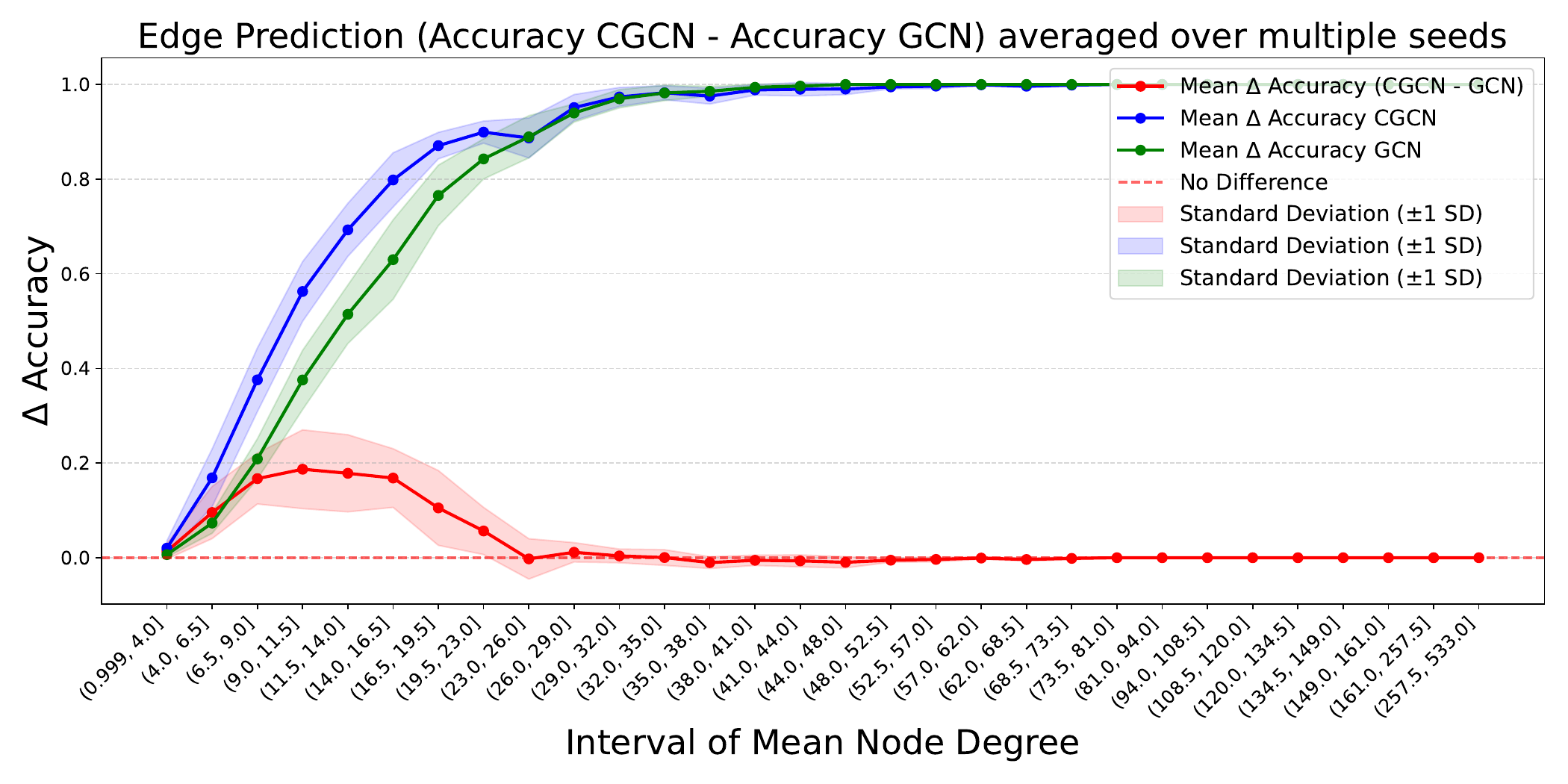}
        \caption{HuRI-PPI}
        \label{fig:deg-huri}
    \end{subfigure}
    \hfill
    \begin{subfigure}[b]{0.45\textwidth}
        \centering
        \includegraphics[width=\textwidth]{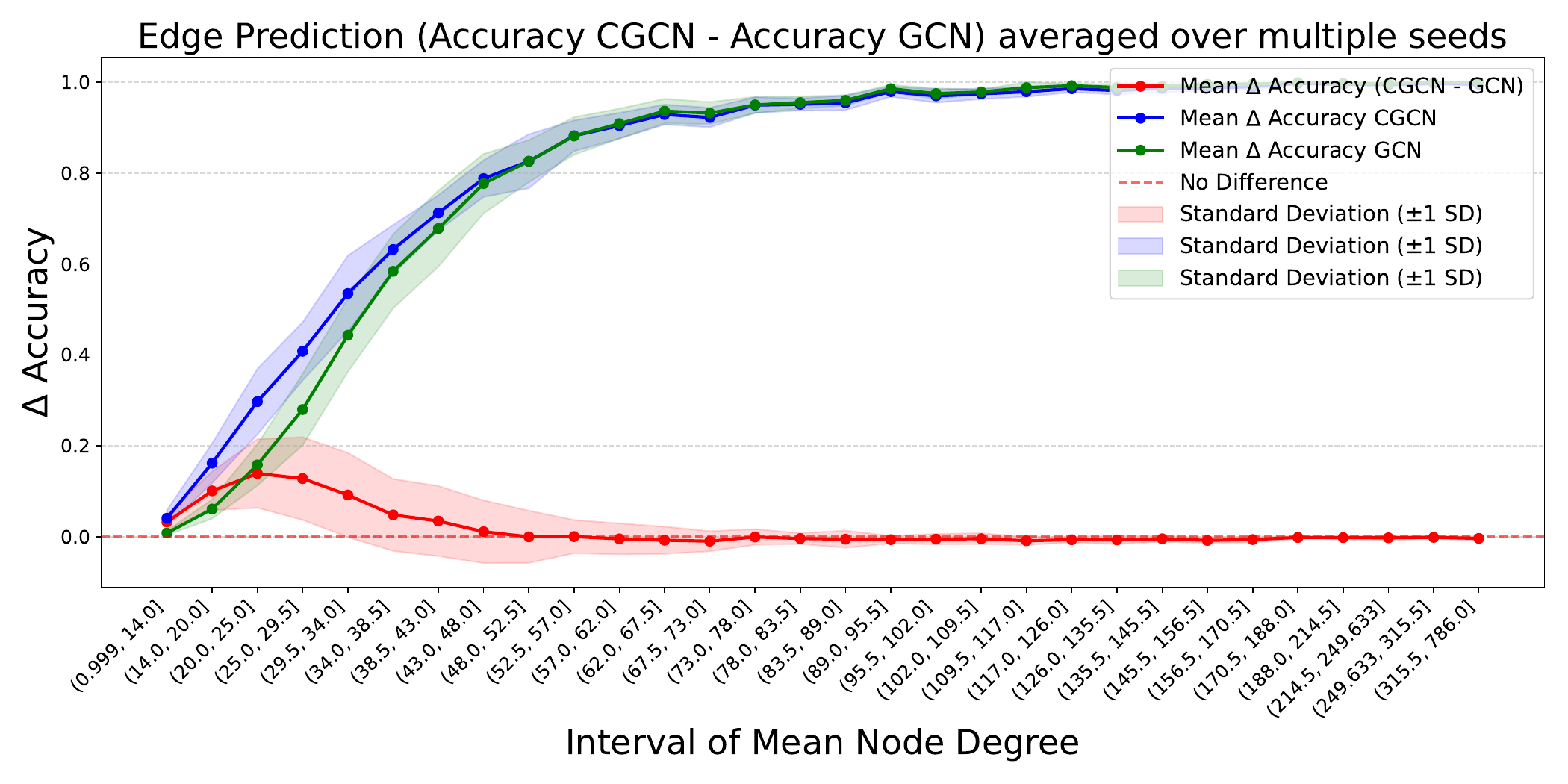}
        \caption{PPT-Ohmnet}
        \label{fig:deg-ohmnet}
    \end{subfigure}

    \vspace{0.5cm}
    \begin{subfigure}[b]{0.45\textwidth}
        \centering
        \includegraphics[width=\textwidth]{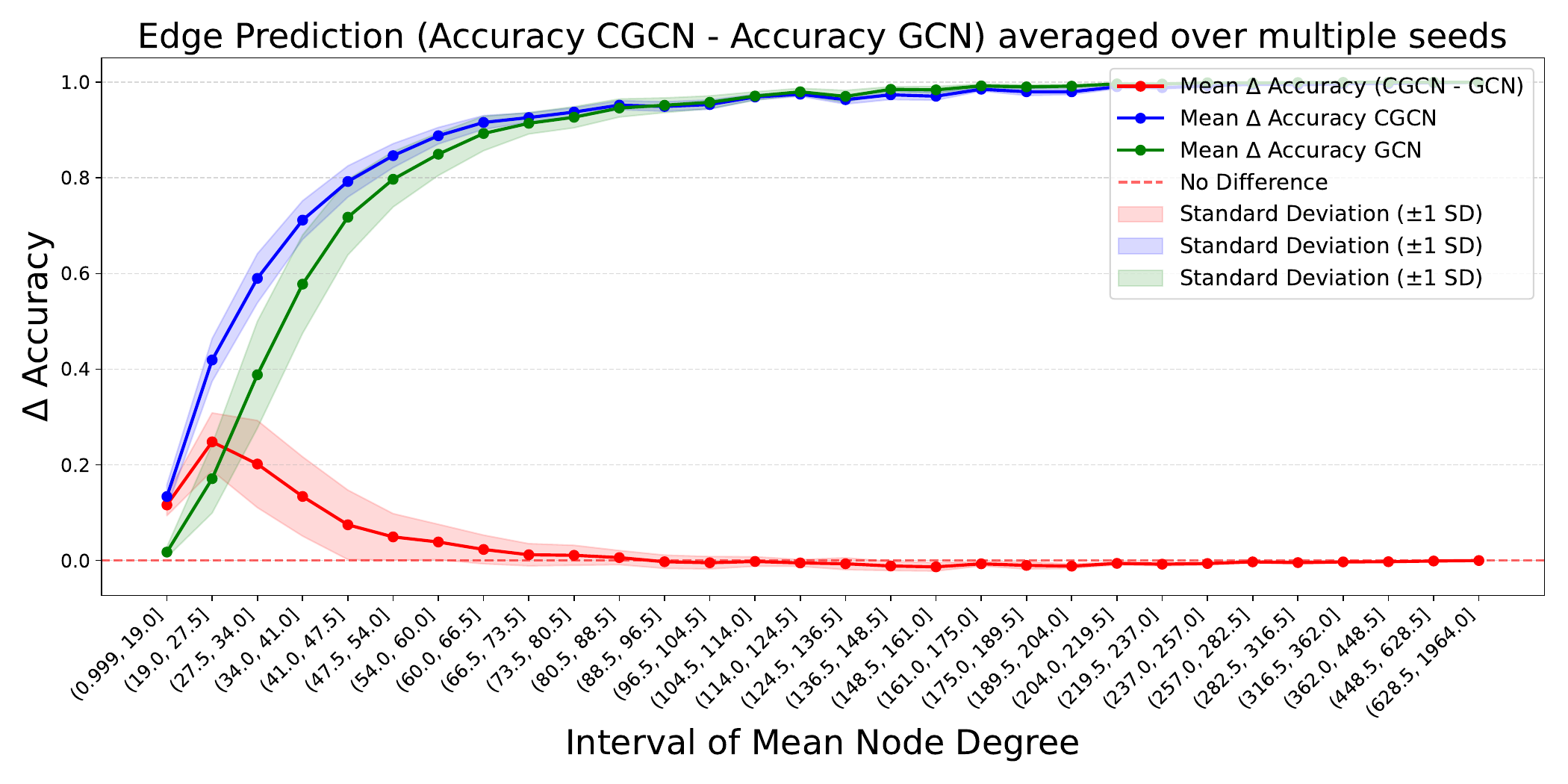}
        \caption{PP-Pathways}
        \label{fig:deg-pathways}
    \end{subfigure}
    \hfill
    \begin{subfigure}[b]{0.45\textwidth}
        \centering
        \includegraphics[width=\textwidth]{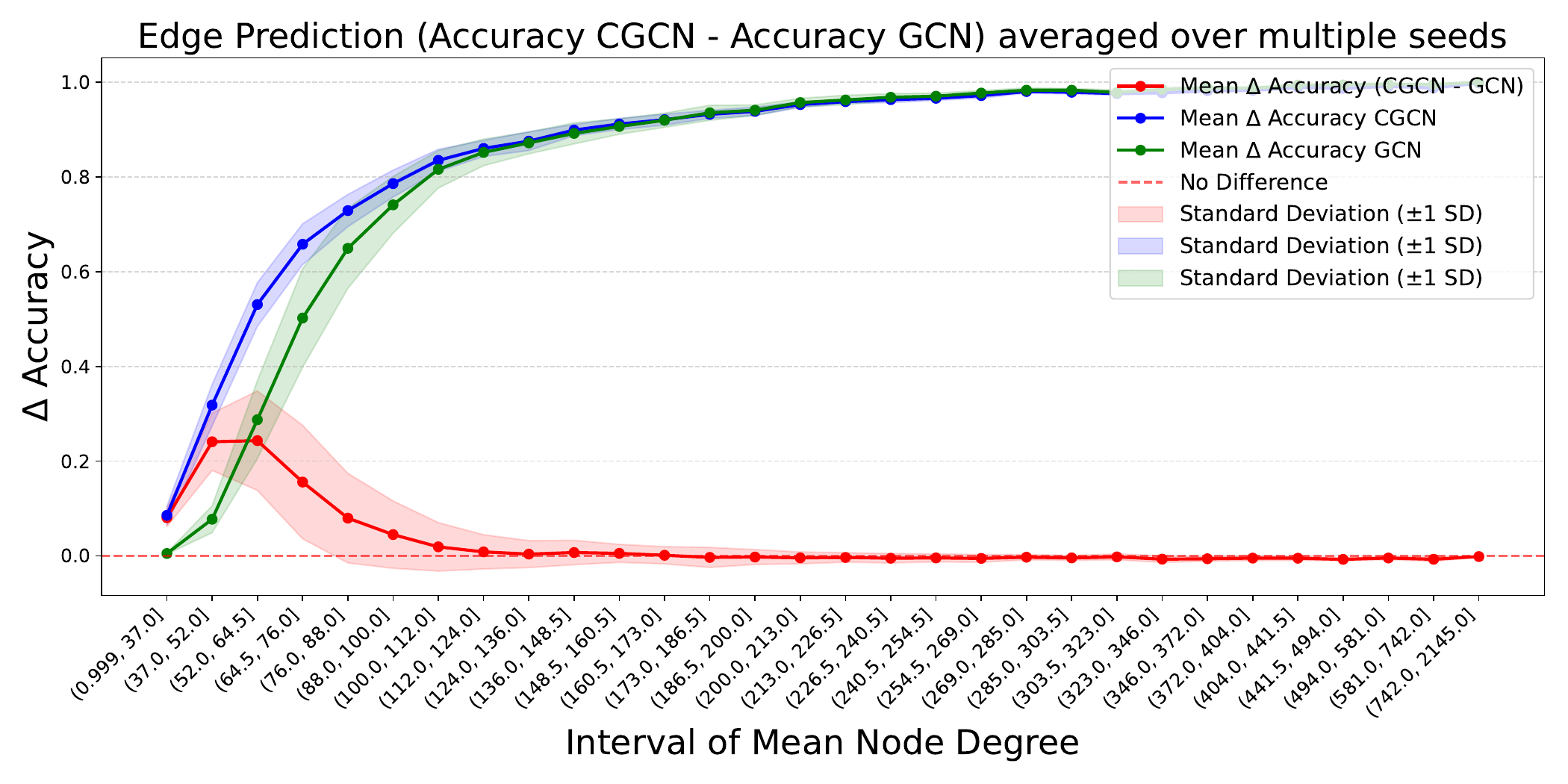}
        \caption{PP-Decagon}
        \label{fig:deg-decagon}
    \end{subfigure}
    \caption{Accuracy difference (CGCN $-$ GCN) as a function of mean node degree, averaged over
    10 random seeds (mean and standard deviation shown), across the four PPI datasets.}
    \label{fig:deg}
\end{figure}

To explain why the improvement is concentrated in low-degree edges, we examine overall
accuracy as a function of mean node degree (Figure~\ref{fig:deg}). For high-degree edges,
GCN's accuracy is already close to 1.0 and CGCN shows little further improvement, suggesting
performance is already near saturation with little room for gains. Low-degree edges, by
contrast, are harder for both models: since node representations are built purely from
neighborhood aggregation with no features beyond graph topology, low-degree nodes yield
representations that carry comparatively little information. This suggests that low-degree
nodes benefit disproportionately from the additional geometric information encoded by the
ORC-dependent weight function.

\begin{figure}[ht]
    \centering
    \begin{subfigure}[b]{0.45\textwidth}
        \centering
        \includegraphics[width=\textwidth]{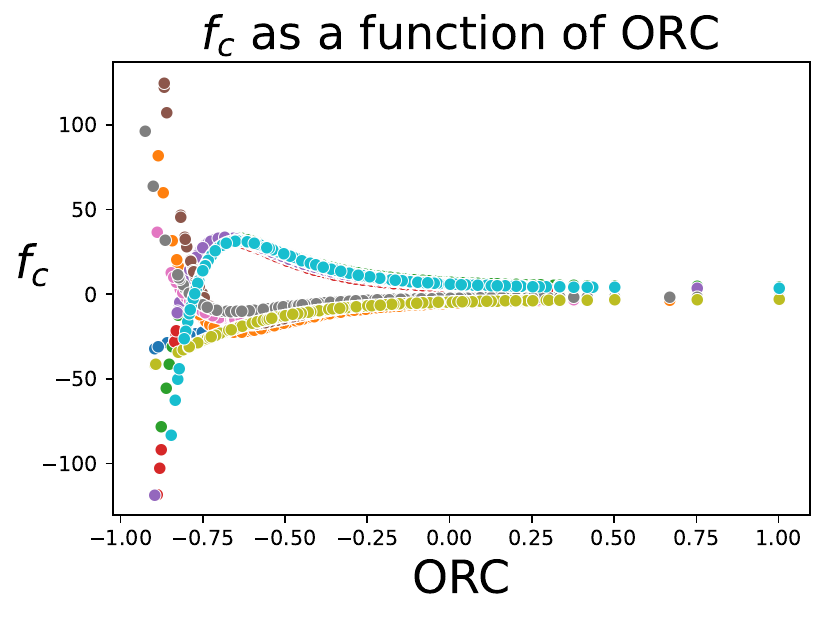}
        \caption{HuRI-PPI}
        \label{fig:forc-huri}
    \end{subfigure}
    \hfill 
    \begin{subfigure}[b]{0.45\textwidth}
        \centering
        \includegraphics[width=\textwidth]{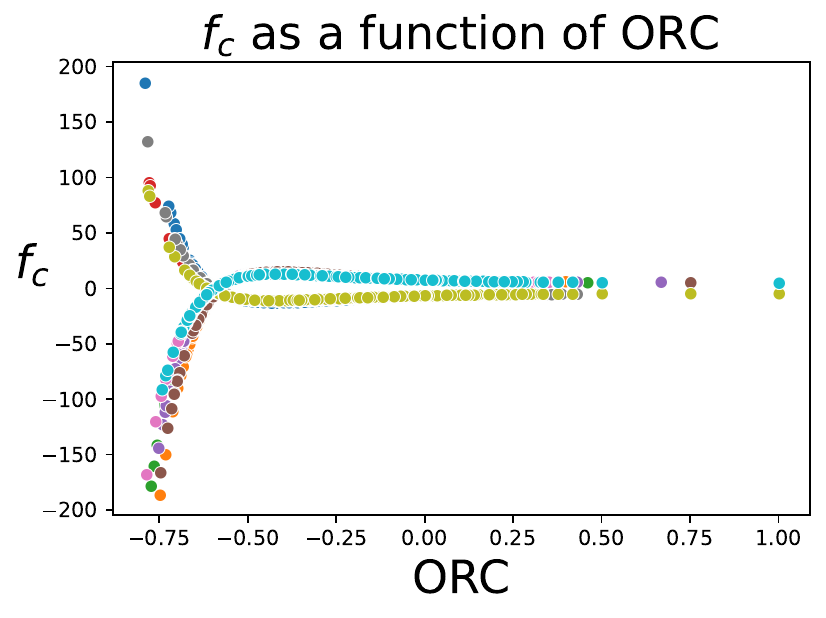}
        \caption{PPT-Ohmnet}
        \label{fig:forc-ohmnet}
    \end{subfigure}
    
    \vspace{0.5cm} 

    \begin{subfigure}[b]{0.45\textwidth}
        \centering
        \includegraphics[width=\textwidth]{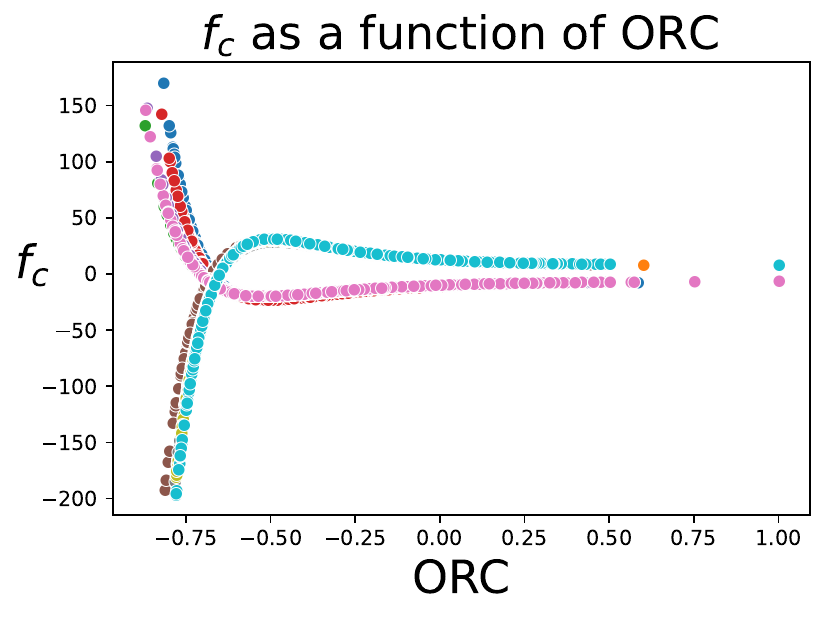}
        \caption{PP-Pathway}
        \label{fig:forc-pathways}
    \end{subfigure}
    \hfill 
    \begin{subfigure}[b]{0.45\textwidth}
        \centering
        \includegraphics[width=\textwidth]{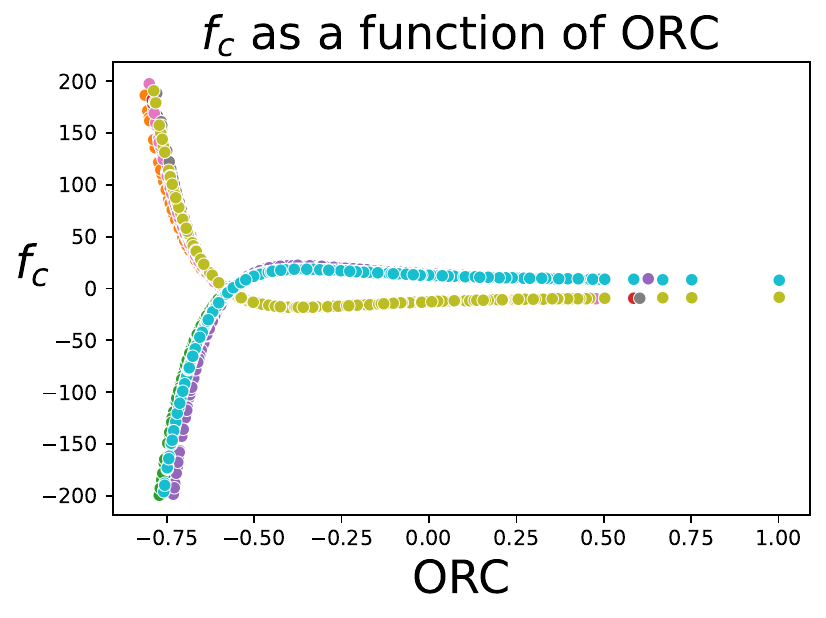}
        \caption{PP-Decagon}
        \label{fig:forc-decagon}
    \end{subfigure}

     \caption{$f$ as a function of ORC for 10 random seeds}
    \label{fig:curva}
\end{figure}

To explain why accuracy drops for low-curvature edges, we first note that HuRI-PPI, obtained
through systematic experimental testing rather than literature curation, is structurally
different from the other three datasets. We therefore focus primarily on PPT-Ohmnet, PP-Decagon, and
PP-Pathways, where the degradation pattern is most consistent. Since $f$ is the only component that differs between GCN and CGCN, the explanation must naturally lie in the behavior of this function. Thus, we analyze $f_c(u,v)$ as a function of $\kappa_{u,v}$.  As $f_{c}$ is learned, it
varies somewhat across runs (Figure~\ref{fig:curva}), but a consistent pattern emerges: $f_{c}$ takes
extreme values at very low curvature, while for curvature values above roughly $-0.5$ it remains
close to zero and of the opposite sign.

This shape suggests that $f_{c}$, used as an edge weight in message passing, allows the model to
specifically distinguish which nodes are reachable from the current node mainly through very-low-curvature (bottleneck) edges. We offer the
following intuition for how this could operate: suppose  $f_{c}$ takes strongly negative values for
low-curvature edges and mildly positive values otherwise (the opposite case is analogous). We now explain how the embedding $x_i^1$ of node $v_i$ is obtained in a 1-layer network. Recall Equation~(\ref{eq:gcn-concrete}).
By definition $x_j^0$ corresponds to the $j$-th row of the adjacency matrix, thus, consists only of $0$ and $1$. $W\cdot x_j^0$ is a vector, where each entry is a weighted sum of the neighbors of $v_j$. Without loss of generality, we assume that $W$ consists only of positive entries, otherwise we could shift all values. Depending on each row of $W$, each weighting usually pays special attention to a certain subset of nodes, and so the corresponding entry in $W\cdot x_j^0$ tells us how many of those nodes are in the neighborhood of $v_j$. If the corresponding final entry in $x_i^1$ is highly positive, it means that many nodes of the considered subset are within the 2-hop neighborhood of $v_i$ and reachable without bottleneck edges, thus, they are probably within the same cluster. On the other hand, if the corresponding final entry in $x_i^1$ is negative, it means that some nodes of the considered subset are within the 2-hop neighborhood of $v_i$, but reachable through at least one low-curvature edge not sufficiently offset by high-curvature paths, so these nodes are probably not in the same cluster. 
This would allow the network to represent cluster structure more sharply in
the node embeddings, at the cost of accuracy on the low-curvature (largely inter-cluster) edges
themselves that are sacrificed to achieve it. This reasoning is of course simplified since each row of $W$ not necessarily only pays attention to one clear subset of nodes, but each node in the entire graph receives a different weighting. However, there will be some nodes that are more weighted than other nodes and the general idea explained above still applies.

We further provide evidence for this cluster-separation intuition. First, we compute the Spearman correlation between edge curvature and the Euclidean/cosine distance between endpoint embeddings (Figure~\ref{fig:corr_dist_curv} in
Appendix~\ref{sec:appendix_curv_dist}); this correlation is negative across all four datasets, and consistently more negative, or equal in case of PP-Decagon, under CGCN than GCN, weakly supporting the idea that CGCN increases embedding-space separation between endpoints of low-curvature edges. Second, we evaluate clustering validity indices (Silhouette, Davies-Bouldin, Calinski-Harabasz) on the test-set node  embeddings, using model-independent precomputed communities extracted by the Leiden algorithm \cite{traag2019leiden} as reference labels (Table~\ref{tab:cluster_index} in Appendix~\ref{sec:appendix_cluster_validity}). Developed as an extension of the popular Louvain algorithm, Leiden is a graph-based community detection method that partitions the graph by optimizing modularity while guaranteeing well-connected clusters. CGCN improves over GCN in all metric-dataset combinations, indicating stronger cluster separation within the node embeddings. A paired $t$-test (assuming normally distributed differences) confirms that the improvement is significant ($p<0.05$) for at least two indices per dataset (Table~\ref{tab:clustering_significance} in Appendix~\ref{sec:appendix_cluster_validity}). Third, we compare mean (normalized) distances between inter- and intra-cluster edges, where clusters refer to the precomputed Leiden communities 
 (Table~\ref{tab:inter_intra} in
Appendix~\ref{sec:appendix_inter_intra}): the ratio of inter-to-intra distances is lower under CGCN than GCN in all four datasets, for both Euclidean and cosine distance implying larger distances in the embedding space for the incident nodes of inter-cluster edges (edges between different clusters) than for intra-cluster edges (edges within the same cluster), supporting the cluster-separation hypothesis. Here, a paired $t$-test on the log-transformed ratios (assuming normally distributed log-differences) confirms high significance ($p<0.001$) for all datasets and distance ratios (Table~\ref{tab:ratio_significance} in Appendix~\ref{sec:appendix_inter_intra}).

As a complementary test of the over-squashing account specifically, we compare performance across message-passing depths 1 through 4 (Table~\ref{tab:layers} in Appendix ~\ref{sec:appendix_layers}). If CGCN's benefit came from mitigating over-squashing, we would expect the relative benefit of CGCN to  grow with hop count, but the CGCN/GCN performance ratio remains essentially unchanged across all four depths for all four datasets. Even though, we have no proof that this necessarily contradicts the over-squashing account, it is also not supporting it, leaving it without any strong evidence.

Full clustering-validity indices, inter/intra-cluster distance statistics, and the layer-depth ablation supporting these claims are reported in Appendix~\ref{sec:appendix_cluster_validity}--\ref{sec:appendix_layers}.

\section{Discussion}
\label{sec:discussion}

Our experimental results strongly support our hypothesis that the improvement of CGCN is due to
better cluster separation, with our analysis of the learned edge-weighting function $f_c$ providing a plausible explanation for this
behavior. It is, however, questionable whether this behavior is desirable or merely reflects
overfitting to topological artifacts rather than capturing true underlying biology, as clusters within the PPT-Ohmnet, PP-Pathways, and PP-Decagon datasets
are likely to originate from study bias \cite{HuRI, rolland2014}. Additionally, in many applications
the prediction of low-curvature (bottleneck) edges could be of particular interest. We observe
that although we also see better cluster separation with CGCN for the HuRI-PPI dataset, this does
not lead to a significant performance degradation on low-curvature edges. We only observe a
slight decrease in the correlation between loss and curvature. We hypothesize that because
HuRI-PPI's community structure is comparatively fine-grained and sparse rather than built from a
few large, densely-clustered communities, its low-curvature edges may more often belong to a
star- or tree-shaped local structure instead of bridging two distinct dense clusters, which would
explain why they are not significantly affected by stronger cluster separation. However,
validating this conjecture would require further testing on a broader range of datasets featuring
diverse structural patterns.

Shen et al. \cite{shen2024curvature} motivate CGCN as a method to alleviate over-squashing. While our analysis
provides an alternative explanation for the success of CGCN, determining whether our explanation
aligns with the reduction of over-squashing or represents an independent phenomenon remains
beyond the scope of this work. We observe that the improvement of CGCN over GCN does not change
significantly as the number of layers increases, which is a condition under which over-squashing
typically becomes more relevant, challenging the claim by Shen et al. \cite{shen2024curvature} that CGCN
alleviates over-squashing. Nevertheless, over-squashing is a complex phenomenon, and based on our
results, we cannot make a definitive claim in either direction.
It is up to future work to investigate the connection or opposition of cluster separation and the prevention of over-squashing.

More interesting further investigations would include the same analysis on other models, which use ORC in message passing, such as CurvGN and CGNN, to check if the cluster separation hypothesis generalizes or is specific to CGCN. Moreover, rewiring-methods could be studied under the same question, however, here we would expect not to find sharper cluster separation, since additional connections around bottleneck edges could blur cluster borders.

Notably, discrete Ricci flow for community detection \cite{ni2019community} exhibits a conceptual overlap with our edge-weighting approach, as both dynamically modify edge properties. However, both rely on fundamentally different mechanisms with distinct objectives, and exploring their precise connection remains an interesting direction for future work.

While the cluster separation hypothesis as presented here is a new idea not mentioned by Shen et al. \cite{shen2024curvature}, they study the performance of CGCN in representation learning with a different focus by comparing the projection into 2D spaces by t-SNE of edge embedding vectors, where positive and negative edges are regarded as the two different classes. Their comparison of CGCN with other network embedding models, such as DeepWalk, LINE, Node2Vec and SDNE, show a better separation between those two classes in CGCN. However, this analysis is lacking to include the representation of GCN, which does not strongly differ from the representation of CGCN (see Figure~\ref{fig:tsne}). So the improved separation between positive and negative edges does not originate from the incorporation of ORC.

\section{Limitations}
\label{sec:limitations}

The observations made in this paper are limited to the four analyzed datasets and are based on
empirical results. The same model might perform differently on other datasets featuring alternative
structural patterns or even on the same datasets if different sampling strategies for the training set are used, which could result in a differently learned edge-weighting function. However, since PPT-Ohmnet, PP-Pathways, and PP-Decagon all show the same
results, we believe that our explanation could generalize to other datasets exhibiting similar
structures, which may be the case for many literature-based PPI datasets or other biomedical
networks. Nevertheless, further tests are required to support this claim.

Moreover, most of our analyses used positive samples (i.e., actual interactions). Similar
investigations could be conducted for negative samples, where we would expect the results to also be influenced by the improved cluster separation.

\section{Conclusion}

Incorporating geometric information into deep learning models can yield substantial benefits and is the core objective of geometric deep learning. Discrete Ricci curvature is a promising candidate as it characterizes the over-squashing phenomenon. However, understanding how geometric information influences model behavior on individual link predictions remains crucial. With this work, we take a step toward a better understanding of the influence of ORC in CGCN, by demonstrating empirically on PPI datasets that CGCN utilizes ORC information during message passing to sharpen cluster separation in the embedding space. While this improves aggregate performance and predictions for low-degree nodes, it comes at the cost of degraded performance on low-curvature (bottleneck) edges. 

This insight provides a valuable foundation for deciding when to deploy CGCN versus standard GCN, depending on the given application and which interactions are most important to predict correctly.
Moreover, it provides a range of directions for promising future work, including testing the cluster-separation account for further models and investigating its relationship with or contrast to over-squashing.

{\small
\bibliographystyle{plain}  
\bibliography{name}         

@article{atz2021geometric,
  title={Geometric deep learning on molecular representations},
  author={Atz, Kenneth and Grisoni, Francesca and Schneider, Gisbert},
  journal={Nature Machine Intelligence},
  volume={3},
  number={12},
  pages={1023--1032},
  year={2021},
  publisher={Nature Publishing Group UK London}
}

@article{moll2006biomolecular,
  title={Biomolecular interaction analysis in functional proteomics},
  author={Moll, Daniela and Prinz, Anke and Gesellchen, Frank and Drewianka, Stephan and Zimmermann, Bastian and Herberg, Friedrich W},
  journal={Journal of neural transmission},
  volume={113},
  number={8},
  pages={1015--1032},
  year={2006},
  publisher={Springer}
}

@article{ollivier2009ricci,
  title={{Ricci} curvature of {Markov} chains on metric spaces},
  author={Ollivier, Yann},
  journal={Journal of Functional Analysis},
  volume={256},
  number={3},
  pages={810--864},
  year={2009},
  publisher={Elsevier}
}

@article{sreejith2016forman,
  title={{Forman} curvature for complex networks},
  author={Sreejith, R. P. and Mohanraj, Karthikeyan and Jost, J{\"u}rgen and Saucan, Emil and Samal, Areejit},
  journal={Journal of Statistical Mechanics: Theory and Experiment},
  volume={2016},
  number={6},
  pages={063206},
  year={2016},
  publisher={IOP Publishing and SISSA},
  doi={10.1088/1742-5468/2016/06/063206}
}

@article{wee2021forman,
  title={Forman persistent {Ricci} curvature (FPRC)-based machine learning models for protein--ligand binding affinity prediction},
  author={Wee, JunJie and Xia, Kelin},
  journal={Briefings in Bioinformatics},
  volume={22},
  number={6},
  pages={bbab136},
  year={2021},
  publisher={Oxford University Press}
}

@article{eidi2020edge,
  title={Edge-based analysis of networks: curvatures of graphs and hypergraphs},
  author={Eidi, Marzieh and Farzam, Amirhossein and Leal, Wilmer and Samal, Areejit and Jost, J{\"u}rgen},
  journal={Theory in biosciences},
  volume={139},
  number={4},
  pages={337--348},
  year={2020},
  publisher={Springer}
}

@article{shen2024curvature,
  title={Curvature-enhanced graph convolutional network for biomolecular interaction prediction},
  author={Shen, Cong and Ding, Pingjian and Wee, Junjie and Bi, Jialin and Luo, Jiawei and Xia, Kelin},
  journal={Computational and Structural Biotechnology Journal},
  volume={23},
  pages={1016--1025},
  year={2024},
  publisher={Elsevier}
}

@article{kipf2016semi,
  title={Semi-supervised classification with graph convolutional networks},
  author={Kipf, Thomas N and Welling, Max},
  journal={arXiv preprint arXiv:1609.02907},
  year={2016}
}

@article{velickovic2017graph,
  title={Graph attention networks},
  author={Veli{\v{c}}kovi{\'c}, Petar and Cucurull, Guillem and Casanova, Arantxa and Romero, Adriana and Lio, Pietro and Bengio, Yoshua},
  journal={arXiv preprint arXiv:1710.10903},
  year={2017}
}

@inproceedings{xu2019powerful,
  title={How Powerful are Graph Neural Networks?},
  author={Xu, Keyulu and Hu, Weihua and Leskovec, Jure and Jegelka, Stefanie},
  booktitle={International Conference on Learning Representations (ICLR)},
  year={2019}
}

@inproceedings{karhadkar2023fosr,
  title={{F}o{SR}: First-order spectral rewiring for addressing oversquashing in {GNN}s},
  author={Karhadkar, Kedar and Banerjee, Pradeep Kr. and Mont{\'u}far, Guido},
  booktitle={International Conference on Learning Representations (ICLR)},
  year={2023}
}

@article{HuRI,
  title={A reference map of the human binary protein interactome},
  author={Luck, Katja and Kim, Dae-Kyum and Lambourne, Luke and Spirohn, Kerstin and Begg, Bridget E and Bian, Wenting and Brignall, Ruth and Cafarelli, Tiziana and Campos-Laborie, Francisco J and Charloteaux, Benoit and others},
  journal={Nature},
  volume={580},
  number={7803},
  pages={402--408},
  year={2020},
  publisher={Nature Publishing Group UK London}
}

@misc{PP,
  author={Zitnik, Marinka and Sosi{\v{c}}, Rok and Maheshwari, Sagar and Leskovec, Jure},
  title={{BioSNAP Datasets}: {Stanford} Biomedical Network Dataset Collection},
  howpublished={\url{http://snap.stanford.edu/biodata}},
  month=aug,
  year={2018}
}

@article{PP-Decagon,
  title={Modeling polypharmacy side effects with graph convolutional networks},
  author={Zitnik, Marinka and Agrawal, Monica and Leskovec, Jure},
  journal={Bioinformatics},
  volume={34},
  number={13},
  pages={i457--i466},
  year={2018},
  publisher={Oxford University Press},
  doi={10.1093/bioinformatics/bty294}
}

@inproceedings{PP-Pathways,
  title={Large-scale analysis of disease pathways in the human interactome},
  author={Agrawal, Monica and Zitnik, Marinka and Leskovec, Jure},
  booktitle={Pacific Symposium on Biocomputing},
  volume={23},
  pages={111--122},
  year={2018}
}

@article{PPT-Ohmnet,
  title={Predicting multicellular function through multi-layer tissue networks},
  author={Zitnik, Marinka and Leskovec, Jure},
  journal={Bioinformatics},
  volume={33},
  number={14},
  pages={i190--i198},
  year={2017},
  publisher={Oxford University Press},
  doi={10.1093/bioinformatics/btx252}
}

@article{rolland2014,
  title={A proteome-scale map of the human interactome network},
  author={Rolland, Thomas and Ta{\c{s}}an, Murat and Charloteaux, Benoit and Pevzner, Samuel J and Zhong, Quan and Sahni, Nidhi and Yi, Song and Lemmens, Irma and Fontanillo, Celia and Mosca, Roberto and others},
  journal={Cell},
  volume={159},
  number={5},
  pages={1212--1226},
  year={2014},
  publisher={Elsevier},
  doi={10.1016/j.cell.2014.10.050}
}

@inproceedings{topping2021understanding,
  title={Understanding over-squashing and bottlenecks on graphs via curvature},
  author={Topping, Jake and Di Giovanni, Francesco and Chamberlain, Benjamin Paul and Dong, Xiaowen and Bronstein, Michael M.},
  booktitle={International Conference on Learning Representations (ICLR)},
  year={2022}
}

@inproceedings{ye2020curvature,
  title={Curvature Graph Network},
  author={Ye, Ze and Liu, Kin Sum and Ma, Tengfei and Gao, Jie and Chen, Chao},
  booktitle={International Conference on Learning Representations (ICLR)},
  year={2020}
}

@article{li2022curvature,
  title={Curvature graph neural network},
  author={Li, Haifeng and Cao, Jun and Zhu, Jiawei and Liu, Yu and Zhu, Qing and Wu, Guohua},
  journal={Information Sciences},
  volume={592},
  pages={50--66},
  year={2022}
}

@article{yadav2024roadmap,
  title={A roadmap for curvature-based geometric data analysis and learning},
  author={Yadav, Yasharth and Xia, Kelin},
  journal={arXiv preprint arXiv:2510.22599},
  year={2025}
}

@inproceedings{chen2020measuring,
  title     = {Measuring and Relieving the Over-Smoothing Problem for Graph Neural Networks from the Topological View},
  author    = {Chen, Deli and Lin, Yankai and Li, Wei and Li, Peng and Zhou, Jie and Sun, Xu},
  booktitle = {Proceedings of the AAAI Conference on Artificial Intelligence},
  volume    = {34},
  number    = {04},
  pages     = {3438--3445},
  year      = {2020}
}

@book{riemann2016hypotheses,
  author    = {Riemann, Bernhard},
  title     = {On the Hypotheses Which Lie at the Bases of Geometry},
  editor    = {Pesic, Peter},
  publisher = {Birkh{\"a}user},
  year      = {2016},
  doi       = {10.1007/978-3-319-26042-6}
}

@article{samal2018comparative,
  title   = {Comparative analysis of two discretizations of Ricci curvature for complex networks},
  author  = {Samal, Areejit and Sreejith, R. P. and Gu, Jiao and Liu, Shiping and Saucan, Emil and Jost, J{\"u}rgen},
  journal = {Scientific Reports},
  volume  = {8},
  number  = {1},
  pages   = {8650},
  year    = {2018},
  publisher = {Nature Publishing Group}
}

@article{traag2019leiden,
  title     = {From Louvain to Leiden: guaranteeing well-connected communities},
  author    = {Traag, Vincent A and Waltman, Ludo and van Eck, Nees Jan},
  journal   = {Scientific Reports},
  volume    = {9},
  number    = {1},
  pages     = {5233},
  year      = {2019},
  publisher = {Nature Publishing Group},
  doi       = {10.1038/s41598-019-41695-z}
}

@article{bronstein2021geometric,
  title={Geometric Deep Learning: Grids, Groups, Graphs, Geodesics, and Gauges},
  author={Bronstein, Michael M and Bruna, Joan and Cohen, Taco and Veli{\v{c}}kovi{\'c}, Petar},
  journal={arXiv preprint arXiv:2104.13478},
  year={2021}
}

@article{ni2019community,
  title={Community detection on networks with Ricci flow},
  author={Ni, Chien-Chun and Lin, Yu-Yao and Luo, Feng and Gao, Jie},
  journal={Scientific reports},
  volume={9},
  number={1},
  pages={9984},
  year={2019},
  publisher={Nature Publishing Group UK London}
}

@inproceedings{ni2018network,
  title={Network alignment by discrete ollivier-ricci flow},
  author={Ni, Chien-Chun and Lin, Yu-Yao and Gao, Jie and Gu, Xianfeng},
  booktitle={International symposium on graph drawing and network visualization},
  pages={447--462},
  year={2018},
  organization={Springer}
}
}


\appendix

\section{Technical appendices and supplementary material}

\subsection{Dataset Details}
\label{app:datasets}
Dataset sourcing and the structural distinction between HuRI-PPI and the other three datasets are
discussed in Section~\ref{dataset}. Table~\ref{tab:metrics} summarizes basic properties of
these four datasets.

\begin{table}[ht]
  \centering
  \begin{tabular}{lcccc}
    \hline
    Metric & HuRI-PPI & PPT-Ohmnet & PP-Pathways & PP-Decagon \\
    \hline
    Nodes & 8272 & 4510 & 21557 & 19081 \\
    Edges & 52548 & 70338 & 342352 & 715612 \\
    Mean degree & 12.71 & 31.19 & 31.76 & 75.01 \\
    Density & 0.15\% & 0.69\% & 0.15\% & 0.39\% \\
    Ratio (pos./neg. curvature) & 0.1 & 0.2 & 0.19 & 0.5 \\
    \hline
  \end{tabular}
  \caption{Basic properties of datasets}
  \label{tab:metrics}
\end{table}

To verify the structural difference between HuRI-PPI and the other three datasets, we report
additional clustering-structure metrics in Table~\ref{tab:cluster_structure}: the average
clustering coefficient (a local measure), transitivity (a global measure), Louvain modularity (a
global measure of cluster separation), and the number of communities detected. HuRI-PPI shows a
substantially lower average clustering coefficient and transitivity than the other three datasets,
alongside a much higher community count. Consequently, HuRI-PPI consists of pronounced
hub-and-spoke structures (star topologies): the neighbors of central proteins rarely interact with
one another, resulting in an almost complete absence of closed triangles.
\begin{table}[ht]
  \centering
  \begin{tabular}{lcccc}
    \hline
    Metric & HuRI-PPI & PPT-Ohmnet & PP-Pathways & PP-Decagon \\
    \hline
    Average clustering coefficient & 0.0681 & 0.1678 & 0.1280 & 0.2335 \\
    Transitivity & 0.0398 & 0.1079 & 0.1189 & 0.2114 \\
    Modularity (Louvain) & 0.4668 & 0.3270 & 0.3942 & 0.4596 \\
    Number of communities & 153 & 33 & 51 & 20 \\
    \hline
  \end{tabular}
  \caption{Clustering-structure metrics for the four PPI datasets. Average clustering coefficient
  is a local measure; transitivity and modularity are global measures, with modularity reflecting
  cluster separation.}
  \label{tab:cluster_structure}
\end{table}

\subsection{Baseline Performance}
\label{app:baseline}

Table~\ref{tab:performance} reports aggregate edge prediction performance of GCN and CGCN across
the four PPI datasets, averaged over 10 random seeds. These results confirm the aggregate
improvement of CGCN over GCN reported by \cite{shen2024curvature}, and serve as the starting
point for the edge-level analysis presented in the main text.

\begin{table}[ht]
  \centering
  \begin{tabular}{lcccc}
    \hline
    Metric & HuRI-PPI & PPT-Ohmnet & PP-Pathways & PP-Decagon \\
    \hline
    Loss & 0.3698 ± 0.0047 & 0.4124 ± 0.0020 & 0.3206 ± 0.0021 & 0.3762 ± 0.0018 \\
    AUC & 0.9174 ± 0.0020 & 0.8943 ± 0.0011 & 0.9388 ± 0.0006 & 0.9144 ± 0.0006 \\
    AUPR & 0.9199 ± 0.0025 & 0.8994 ± 0.0012 & 0.9369 ± 0.0006 & 0.9132 ± 0.0008 \\
    F1 & 0.8511 ± 0.0037 & 0.8152 ± 0.0046 & 0.8652 ± 0.0039 & 0.8345 ± 0.0036 \\
    \hline
    Loss & 0.3366 ± 0.0062 & 0.3833 ± 0.0060 & 0.2876 ± 0.0044 & 0.3151 ± 0.0040 \\
    AUC & 0.9313 ± 0.0031 & 0.9101 ± 0.0032 & 0.9506 ± 0.0017 & 0.9407 ± 0.0014 \\
    AUPR & 0.9326 ± 0.0031 & 0.9138 ± 0.0032 & 0.9487 ± 0.0023 & 0.9417 ± 0.0016 \\
    F1 & 0.8632 ± 0.0024 & 0.8308 ± 0.0045 & 0.8837 ± 0.0023 & 0.8675 ± 0.0023 \\
    \hline
  \end{tabular}
  \caption{Edge prediction performance of GCN and CGCN across four PPI datasets, averaged over
  10 random seeds (mean $\pm$ standard deviation).}
  \label{tab:performance}
\end{table}

\subsection{Curvature and Embedding Distance}
\label{sec:appendix_curv_dist}
\begin{figure}[ht]
    \centering
    \begin{subfigure}[b]{0.45\textwidth}
        \centering
        \includegraphics[width=\textwidth]{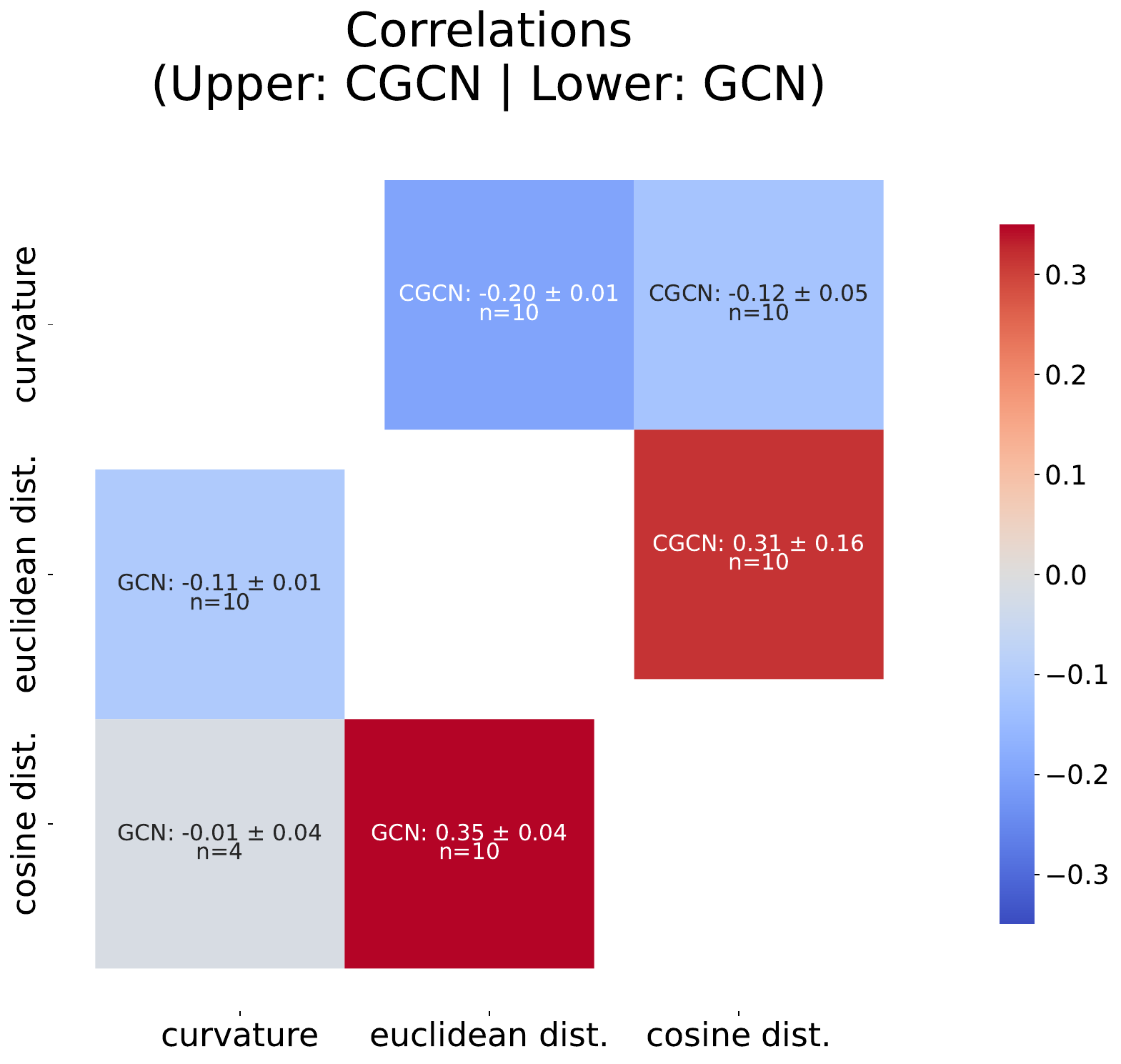}
        \caption{HuRI-PPI}
        \label{fig:bild11}
    \end{subfigure}
    \hfill
    \begin{subfigure}[b]{0.45\textwidth}
        \centering
        \includegraphics[width=\textwidth]{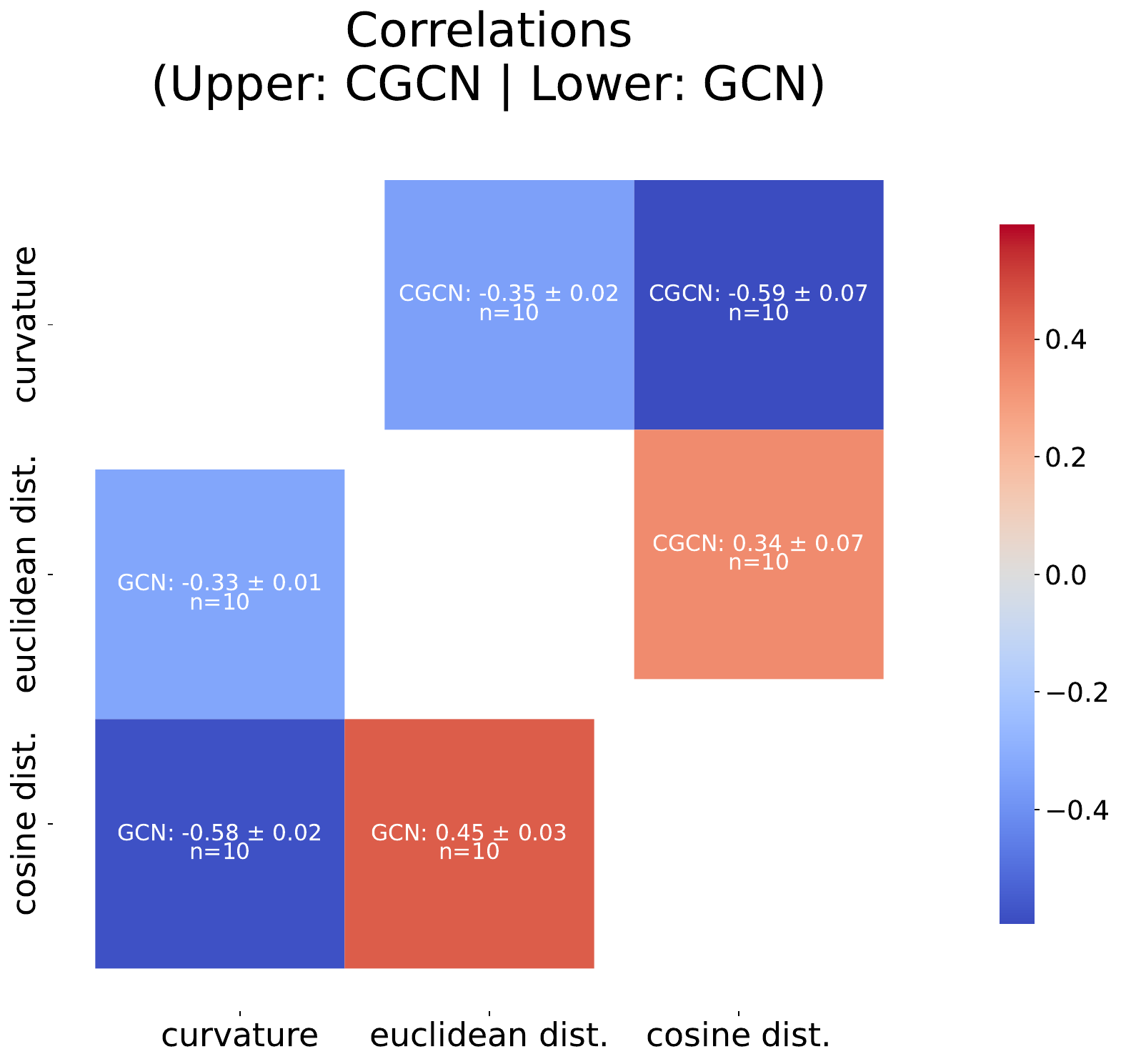}
        \caption{PPT-Ohmnet}
        \label{fig:bild21}
    \end{subfigure}

    \vspace{0.5cm}
    \begin{subfigure}[b]{0.45\textwidth}
        \centering
        \includegraphics[width=\textwidth]{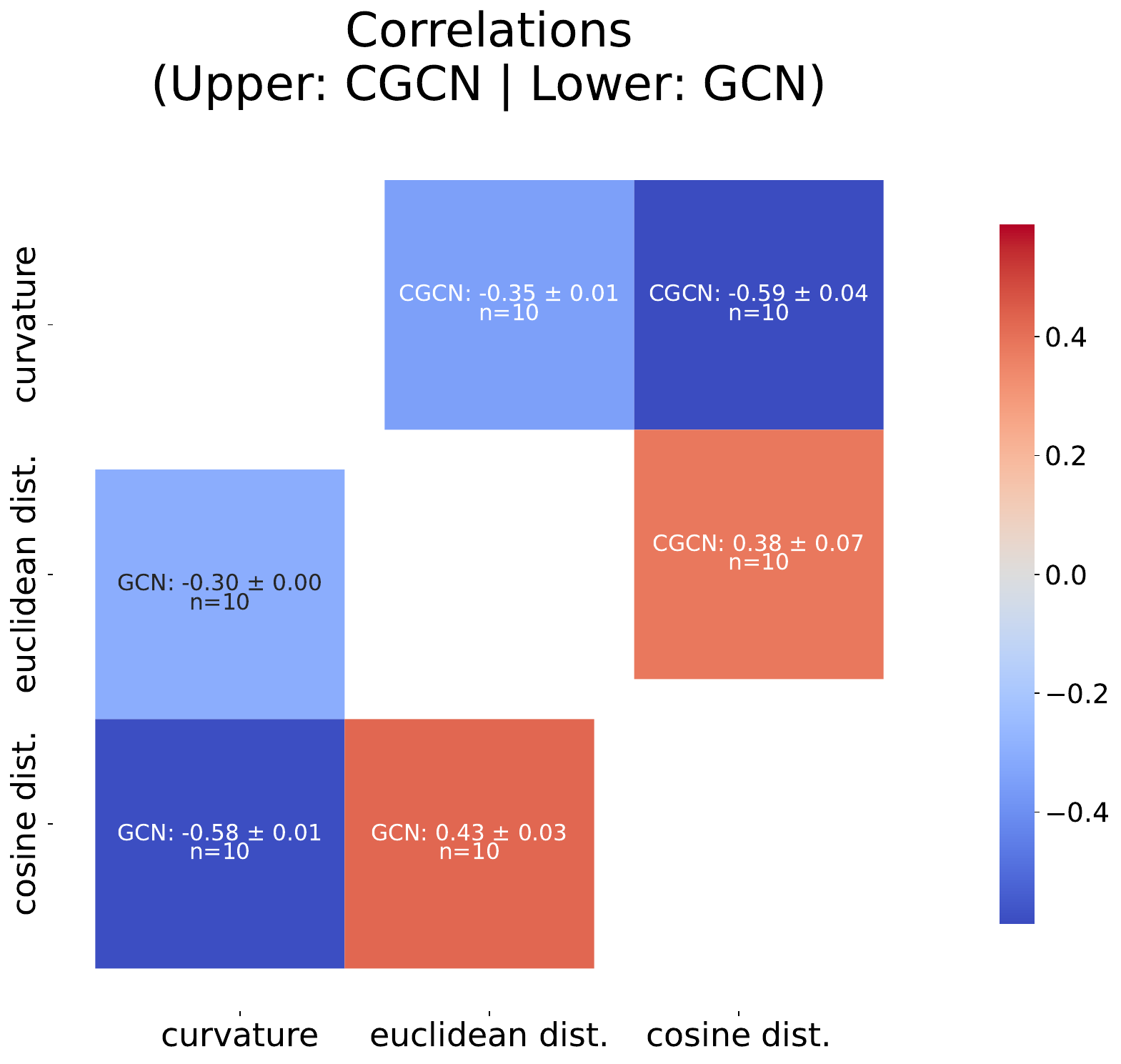}
        \caption{PP-Pathways}
        \label{fig:bild31}
    \end{subfigure}
    \hfill
    \begin{subfigure}[b]{0.45\textwidth}
        \centering
        \includegraphics[width=\textwidth]{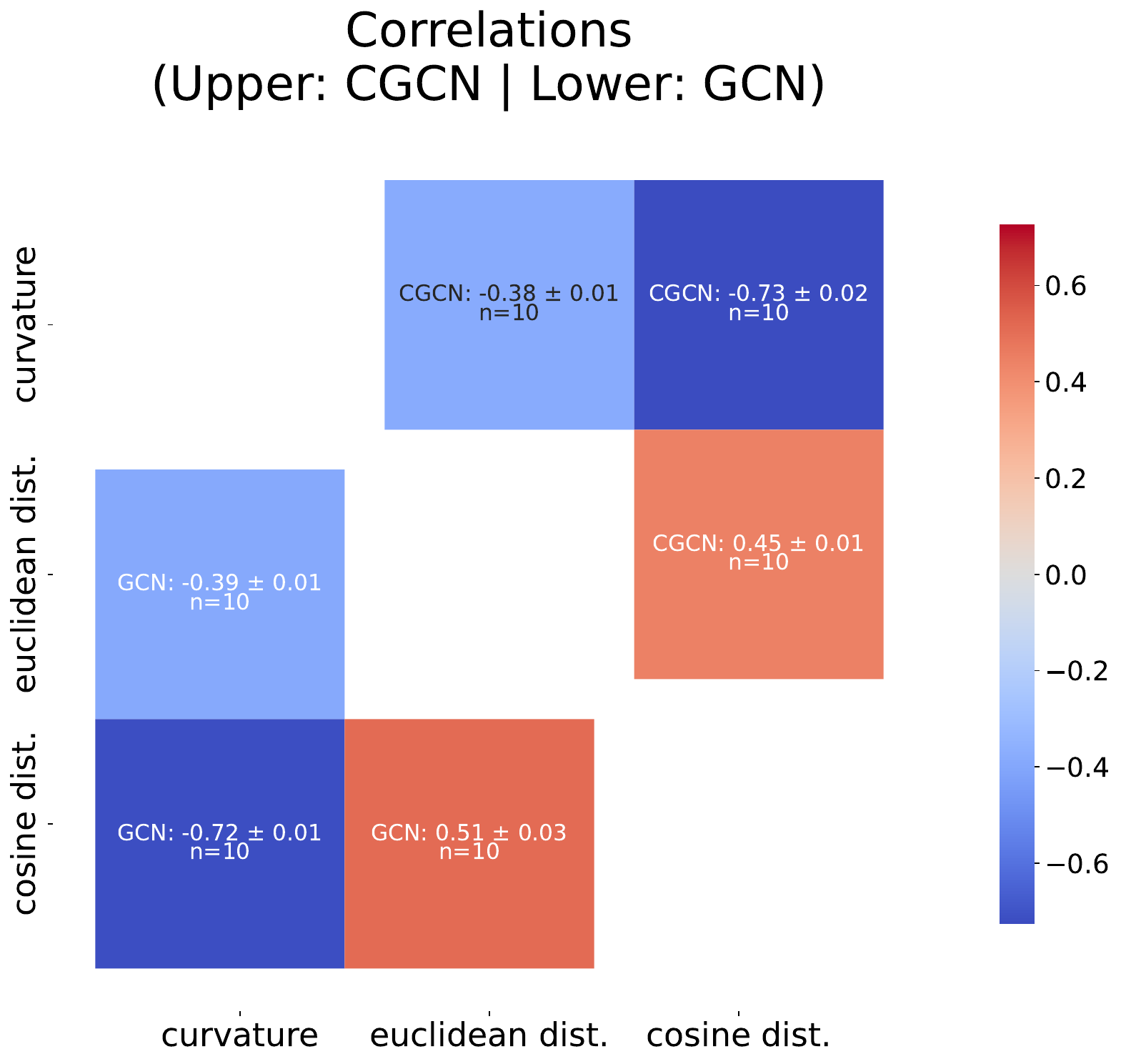}
        \caption{PP-Decagon}
        \label{fig:bild41}
    \end{subfigure}
     \caption{Correlation between edge curvature and Euclidean/cosine distance between endpoint embeddings, for GCN and CGCN, across the four PPI datasets, averaged over 10 random seeds (mean $\pm$ standard deviation).}
    \label{fig:corr_dist_curv}
\end{figure}

To further probe the cluster-separation account, we compute the Spearman correlation between edge curvature and the Euclidean and cosine distance between endpoint embeddings, for both CGCN and GCN (Figure~\ref{fig:corr_dist_curv}). For all datasets, we observe a negative correlation between curvature and Euclidean/cosine distance: lower-curvature (more bottleneck-like) edges tend to have more distant endpoints in embedding space, while higher-curvature edges have closer endpoints. This correlation is either slightly more negative under CGCN than GCN, or comparable between the two. This direction is consistent with the cluster-separation account, though the effect size is modest and this result alone does not strongly support it.

\subsection{Clustering-Validity Indices}
\label{sec:appendix_cluster_validity}

We evaluate clustering-validity indices on the learned node  embeddings (using normalized
Euclidean distances) to assess how well GCN and CGCN separate model-independent precomputed Leiden communities
in embedding space. Silhouette Score (range $[-1,1]$; larger is better
cluster separation), Davies-Bouldin Index (range $[0,\infty)$; smaller is better), and
Calinski-Harabasz Index (range $[0,\infty)$; larger is better) are reported in
Table~\ref{tab:cluster_index}. CGCN shows better values than GCN in all
metric–dataset combinations.

\begin{table}[ht]
  \centering
  \footnotesize
  \setlength{\tabcolsep}{4pt}
  \renewcommand{\arraystretch}{0.95}
  \begin{tabular}{llcccc}
    \hline
    & Index & HuRI-PPI & PPT-Ohmnet & PP-Pathways & PP-Decagon \\
    \hline
    \multirow{3}{*}{\rotatebox[origin=c]{90}{GCN}}
    & Silhouette Score & -0.3259 $\pm$ 0.0374 & -0.2704 $\pm$ 0.0161 & -0.2943 $\pm$ 0.0430 & -0.2892 $\pm$ 0.0230 \\
    & Davies-Bouldin Index & 6.5576 $\pm$ 0.9704 & 9.4410 $\pm$ 1.6368 & 8.3551 $\pm$ 1.0961 & 8.2036 $\pm$ 1.0631 \\
    & Calinski-Harabasz Index & 5.0689 $\pm$ 0.3706 & 10.4508 $\pm$ 2.0219 & 31.7995 $\pm$ 2.5994 & 123.4901 $\pm$ 20.1581 \\
    \hline
    \multirow{3}{*}{\rotatebox[origin=c]{90}{CGCN}}
    & Silhouette Score & -0.2884 $\pm$ 0.0445 & -0.1914 $\pm$ 0.0196 & -0.2841 $\pm$ 0.0365 & -0.2384 $\pm$ 0.0269 \\
    & Davies-Bouldin Index & 5.8551 $\pm$ 0.8547 & 7.5045 $\pm$ 1.3844 & 6.5150 $\pm$ 1.7034 & 6.5396 $\pm$ 0.9320 \\
    & Calinski-Harabasz Index & 5.0948 $\pm$ 1.2985 & 29.8631 $\pm$ 14.2271 & 78.0663 $\pm$ 26.5887 & 249.1650 $\pm$ 22.4835 \\
    \hline
  \end{tabular}
  \caption{Clustering-validity indices for GCN and CGCN, across the four PPI datasets, averaged over
  10 random seeds (mean $\pm$ standard deviation). Silhouette
  Score: larger is better cluster separation. Davies-Bouldin Index: smaller is better.
  Calinski-Harabasz Index: larger is better.}
  \label{tab:cluster_index}
\end{table}

Additionally, we conduct a paired $t$-test to evaluate whether the clustering-validity indices are significantly improved for CGCN compared to GCN, under the assumption that the paired differences are normally distributed (Table~\ref{tab:clustering_significance}).

\begin{table}[htbp]
\centering
  \footnotesize
  \setlength{\tabcolsep}{4pt}
  \renewcommand{\arraystretch}{0.95}
\begin{tabular}{lcccc}
\toprule
Metric & HuRI-PPI & PPT-Ohmnet & PP-Pathways & PP-Decagon \\
\midrule
Silhouette Score     & 0.0152*  & $< 0.0001$*** & 0.2775\phantom{***}  & 0.0002*** \\
Davies-Bouldin Index & 0.0299*  & 0.0001***     & 0.0013**\phantom{*}  & 0.0019**\phantom{*}  \\
Calinski-Harabasz    & 0.4785\phantom{* } & 0.0007***     & 0.0003***            & $< 0.0001$*** \\
\bottomrule
\end{tabular}
\caption{Statistical significance ($p$-values from paired $t$-tests) comparing CGCN against GCN across clustering-validity indices. Directed hypotheses: CGCN $>$ GCN for Silhouette and Calinski-Harabasz; CGCN $<$ GCN for Davies-Bouldin. *:$p < 0.05$, **:$p < 0.01$, ***:$p < 0.001$}
\label{tab:clustering_significance}
\end{table}

\subsection{Inter- vs.\ Intra-Cluster Edge Distances}
\label{sec:appendix_inter_intra}

\begin{table}[ht]
  \centering
  \footnotesize
  \setlength{\tabcolsep}{4pt}
  \renewcommand{\arraystretch}{0.95}
  \begin{tabular}{llcccc}
    \hline
    & Index & HuRI-PPI & PPT-Ohmnet & PP-Pathways & PP-Decagon \\
    \hline
    \multirow{9}{*}{\rotatebox[origin=c]{90}{GCN}}
    & Inter: mean eucl. dist. & 0.7353 $\pm$ 0.0503 & 0.7380 $\pm$ 0.0333 & 0.5347 $\pm$ 0.0641 & 0.4217 $\pm$ 0.0259 \\
    & Intra: mean eucl. dist. & 0.6279 $\pm$ 0.0676 & 0.7993 $\pm$ 0.0351 & 0.5806 $\pm$ 0.0718 & 0.5648 $\pm$ 0.0311 \\
    & Ratio (Inter/Intra): mean eucl. dist. & 1.1766 $\pm$ 0.0575 & 0.9234 $\pm$ 0.0165 & 0.9215 $\pm$ 0.0141 & 0.7464 $\pm$ 0.0104 \\
    \hdashline
    & Inter: mean cos. dist. & 0.3675 $\pm$ 0.0338 & 0.3648 $\pm$ 0.0267 & 0.2331 $\pm$ 0.0467 & 0.1583 $\pm$ 0.0148 \\
    & Intra: mean cos. dist. & 0.2544 $\pm$ 0.0451 & 0.3977 $\pm$ 0.0291 & 0.2349 $\pm$ 0.0523 & 0.2248 $\pm$ 0.0211 \\
    & Ratio (Inter/Intra): mean cos. dist. & 1.4695 $\pm$ 0.1493 & 0.9174 $\pm$ 0.0237 & 0.9979 $\pm$ 0.0325 & 0.7045 $\pm$ 0.0137 \\
    \hdashline
    & Inter: ORC & -0.1440 $\pm$ 0.0026 & -0.0741 $\pm$ 0.0027 & -0.0994 $\pm$ 0.0018 & 0.0106 $\pm$ 0.0033 \\
    & Intra: ORC & -0.3311 $\pm$ 0.0079 & -0.1548 $\pm$ 0.0034 & -0.2354 $\pm$ 0.0035 & -0.1754 $\pm$ 0.0037 \\
    & Ratio (Inter/Intra): ORC & 0.4350 $\pm$ 0.0106 & 0.4788 $\pm$ 0.0223 & 0.4224 $\pm$ 0.0127 & -0.0603 $\pm$ 0.0194 \\
    \hline
    \multirow{6}{*}{\rotatebox[origin=c]{90}{CGCN}}
    & Inter: mean eucl. dist. & 0.8727 $\pm$ 0.0620 & 0.7522 $\pm$ 0.0940 & 0.6175 $\pm$ 0.0695 & 0.5719 $\pm$ 0.0215 \\
    & Intra: mean eucl. dist. & 0.8822 $\pm$ 0.0613 & 0.8922 $\pm$ 0.0574 & 0.7450 $\pm$ 0.0562 & 0.8300 $\pm$ 0.0226 \\
    & Ratio (Inter/Intra): mean eucl. dist. & 0.9899 $\pm$ 0.0403 & 0.8400 $\pm$ 0.0546 & 0.8268 $\pm$ 0.0402 & 0.6891 $\pm$ 0.0178 \\
    \hdashline
    & Inter: mean cos. dist. & 0.4388 $\pm$ 0.0601 & 0.3591 $\pm$ 0.0821 & 0.2583 $\pm$ 0.0571 & 0.2363 $\pm$ 0.0177 \\
    & Intra: mean cos. dist. & 0.4257 $\pm$ 0.0531 & 0.4497 $\pm$ 0.0591 & 0.3223 $\pm$ 0.0491 & 0.4013 $\pm$ 0.0212 \\
    & Ratio (Inter/Intra): mean cos. dist. & 1.0323 $\pm$ 0.0862 & 0.7890 $\pm$ 0.0807 & 0.7941 $\pm$ 0.0688 & 0.5886 $\pm$ 0.0281 \\
    \hline
  \end{tabular}
  \caption{Inter- vs.\ intra-cluster edge distances and ORC, for GCN and CGCN, across the four PPI datasets, averaged over
  10 random seeds (mean $\pm$ standard deviation). ORC is an intrinsic graph property and is identical for both models (see text); it is reported once, in the GCN block.}
  \label{tab:inter_intra}
\end{table}

Inter- vs.\ intra-cluster edge distances are reported in Table~\ref{tab:inter_intra}. Here, clusters refer to the model-independent precomputed Leiden communities.
 Euclidean distances are normalized. The inter/intra
distance ratio is smaller under CGCN than GCN in all four datasets, for both Euclidean and cosine
distance, indicating stronger separation between clusters, represented through a larger distance in embedding space.  ORC itself is an intrinsic property of the graph and is therefore identical for GCN and
CGCN; intra-cluster edges are consistently more negatively curved than inter-cluster edges across
all four datasets.

Additionally, we conduct a paired t-test on the log-transformed ratios (to account for the skewness of ratio distributions and ensure valid normality assumptions) to evaluate whether the ratios are significantly smaller for CGCN compared to GCN (Table~\ref{tab:ratio_significance})."

\begin{table}[htbp]
\centering
\footnotesize
\setlength{\tabcolsep}{4pt}
\renewcommand{\arraystretch}{0.95}
\begin{tabular}{lcccc}
\toprule
Metric Ratio & HuRI-PPI & PPT-Ohmnet & PP-Pathways & PP-Decagon \\
\midrule
Euclidean Ratio & $< 0.0001$*** & 0.0003*** &  0.0001*** & $< 0.0001$*** \\
Cosine Ratio    & $< 0.0001$*** & 0.0003*** & $< 0.0001$*** & $< 0.0001$*** \\
\bottomrule
\multicolumn{5}{l}{\small ***$p < 0.001$}
\end{tabular}
\caption{Statistical significance ($p$-values from paired $t$-tests) comparing inter-/intra-cluster distance ratios between CGCN and GCN. Directed hypothesis: $\text{Ratio}_{\text{CGCN}} < \text{Ratio}_{\text{GCN}}$.}
\label{tab:ratio_significance}
\end{table}

\subsection{Effect of Message-Passing Depth}
\label{sec:appendix_layers}

As a complementary test of the over-squashing account specifically, we compare performance across message-passing depths 1 through 4 (Table~\ref{tab:layers}); if CGCN's benefit came from mitigating over-squashing, this benefit should grow with hop count, but the CGCN/GCN performance ratio remains essentially unchanged across all four depths for HuRI-PPI and PPT-Ohmnet, which is
leaving the over-squashing account without any supporting evidence.

\begin{table}[ht]
  \centering
  \footnotesize
  \begin{tabular}{lcccc}
    \hline
    Metric & 1 layer & 2 layers & 3 layers & 4 layers \\
    \hline
    \multicolumn{5}{l}{\emph{HuRI-PPI}} \\
    Ratio AUROC (CGCN/GCN) & 1.0152 $\pm$ 0.0026 & 1.0143 $\pm$ 0.0029 & 1.0148 $\pm$ 0.0032 & 1.0148 $\pm$ 0.0032 \\
    Ratio AUPR (CGCN/GCN)  & 1.0137 $\pm$ 0.0028 & 1.0147 $\pm$ 0.0028 & 1.0144 $\pm$ 0.0025 & 1.0146 $\pm$ 0.0025 \\
    Ratio F1 (CGCN/GCN)    & 1.0142 $\pm$ 0.0043 & 1.0149 $\pm$ 0.0047 & 1.0174 $\pm$ 0.0052 & 1.0166 $\pm$ 0.0054 \\
    Ratio Accuracy (CGCN/GCN) & 1.0102 $\pm$ 0.0033 & 1.0114 $\pm$ 0.0038 & 1.0122 $\pm$ 0.0032 & 1.0115 $\pm$ 0.0038 \\
    Ratio Loss (CGCN/GCN)  & 0.9103 $\pm$ 0.0159 & 0.9066 $\pm$ 0.0184 & 0.9031 $\pm$ 0.0168 & 0.9018 $\pm$ 0.0211 \\
    \hline
    \multicolumn{5}{l}{\emph{PPT-Ohmnet}} \\
    Ratio AUROC (CGCN/GCN) & 1.0176 $\pm$ 0.0035 & 1.0178 $\pm$ 0.0036 & 1.0176 $\pm$ 0.0050 & 1.0176 $\pm$ 0.0049 \\
    Ratio AUPR (CGCN/GCN)  & 1.0160 $\pm$ 0.0033 & 1.0162 $\pm$ 0.0035 & 1.0159 $\pm$ 0.0045 & 1.0159 $\pm$ 0.0046 \\
    Ratio F1 (CGCN/GCN)    & 1.0191 $\pm$ 0.0067 & 1.0196 $\pm$ 0.0075 & 1.0182 $\pm$ 0.0061 & 1.0186 $\pm$ 0.0054 \\
    Ratio Accuracy (CGCN/GCN) & 1.0186 $\pm$ 0.0041 & 1.0192 $\pm$ 0.0043 & 1.0187 $\pm$ 0.0049 & 1.0189 $\pm$ 0.0048 \\
    Ratio Loss (CGCN/GCN)  & 0.9295 $\pm$ 0.0170 & 0.9277 $\pm$ 0.0159 & 0.9303 $\pm$ 0.0190 & 0.9310 $\pm$ 0.0188 \\
    \hline
    \multicolumn{5}{l}{\emph{PP-Pathways}} \\
    Ratio AUROC (CGCN/GCN) & 1.0126 $\pm$ 0.0016 & 1.0130 $\pm$ 0.0015 & 1.0130 ± 0.0015 & 1.0131 ± 0.0018 \\
    Ratio AUPR (CGCN/GCN)  & 1.0125 $\pm$ 0.0023 & 1.0133 $\pm$ 0.0022 & 1.0132 ± 0.0021 & 1.0133 ± 0.0025 \\
    Ratio F1 (CGCN/GCN)    & 1.0214 $\pm$ 0.0044 & 1.0204 $\pm$ 0.0045 & 1.0194 ± 0.0036 & 1.0205 ± 0.0034 \\
    Ratio Accuracy (CGCN/GCN) & 1.0199 $\pm$ 0.0029 & 1.0199 $\pm$ 0.0025 & 1.0196 ± 0.0026 & 1.0204 ± 0.0025 \\
    Ratio Loss (CGCN/GCN)  & 0.8972 $\pm$ 0.0122 & 0.8968 $\pm$ 0.0144 & 0.8935 ± 0.0140 & 0.8929 ± 0.0137 \\
    \hline
    \multicolumn{5}{l}{\emph{PP-Decagon}} \\
    Ratio AUROC (CGCN/GCN) & 1.0287 $\pm$ 0.0016 & 1.0281 ± 0.0008 & 1.0281 ± 0.0008 & 1.0284 ± 0.0014 \\
    Ratio AUPR (CGCN/GCN)  & 1.0312 $\pm$ 0.0018 & 1.0307 ± 0.0010 & 1.0306 ± 0.0010 & 1.0310 ± 0.0017 \\
    Ratio F1 (CGCN/GCN)    & 1.0395 $\pm$ 0.0070 &  1.0386 ± 0.0058 & 1.0384 ± 0.0047 &  1.0389 ± 0.0040 \\
    Ratio Accuracy (CGCN/GCN) & 1.0406 $\pm$ 0.0035 &  1.0401 ± 0.0015 & 1.0398 ± 0.0018 & 1.0401 ± 0.0019 \\
    Ratio Loss (CGCN/GCN)  & 0.8377 $\pm$ 0.0135 & 0.8394 ± 0.0060 & 0.8422 ± 0.0058 & 0.8400 ± 0.0092 \\
    \hline
  \end{tabular}
  \caption{Ratio of CGCN to GCN performance (CGCN/GCN) across message-passing depths 1--4,
  for all five metrics, averaged over
  10 random seeds (mean $\pm$ standard deviation). Ratios greater than 1 indicate CGCN outperforms GCN (greater than 1 for
  AUROC/AUPR/F1/Accuracy; less than 1 for Loss, where lower is better).}
  \label{tab:layers}
\end{table}

\subsection{Representation Learning for Binary Edge-Embedding}
\label{app:tsne}
\begin{figure}[htbp]
    \centering
    \begin{subfigure}[b]{0.23\textwidth}
        \centering
        \includegraphics[width=\textwidth]{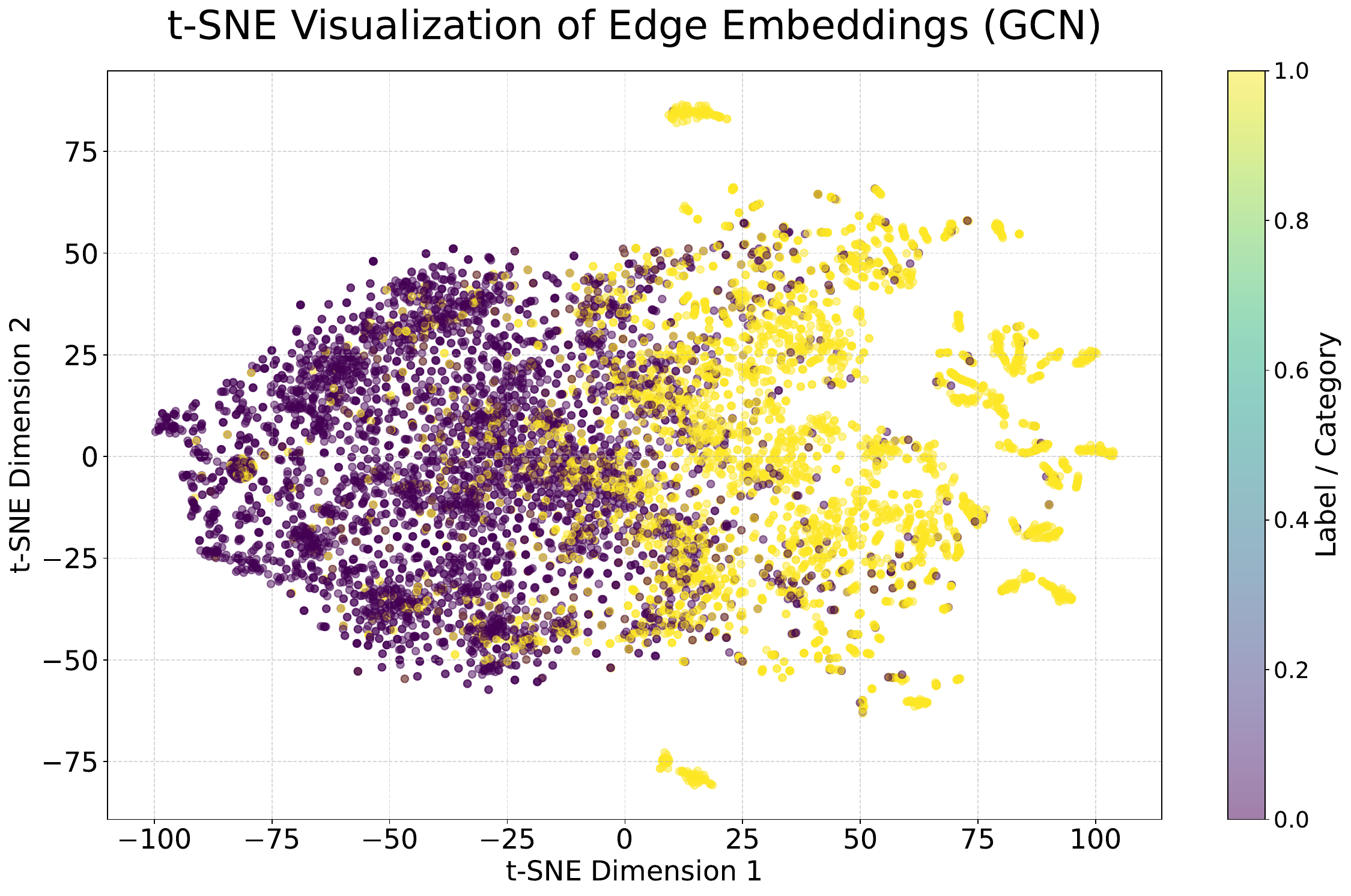}
        \caption{GCN HuRI-PPI}
        \label{fig:bild1_t}
    \end{subfigure}
    \hfill
    \begin{subfigure}[b]{0.23\textwidth}
        \centering
        \includegraphics[width=\textwidth]{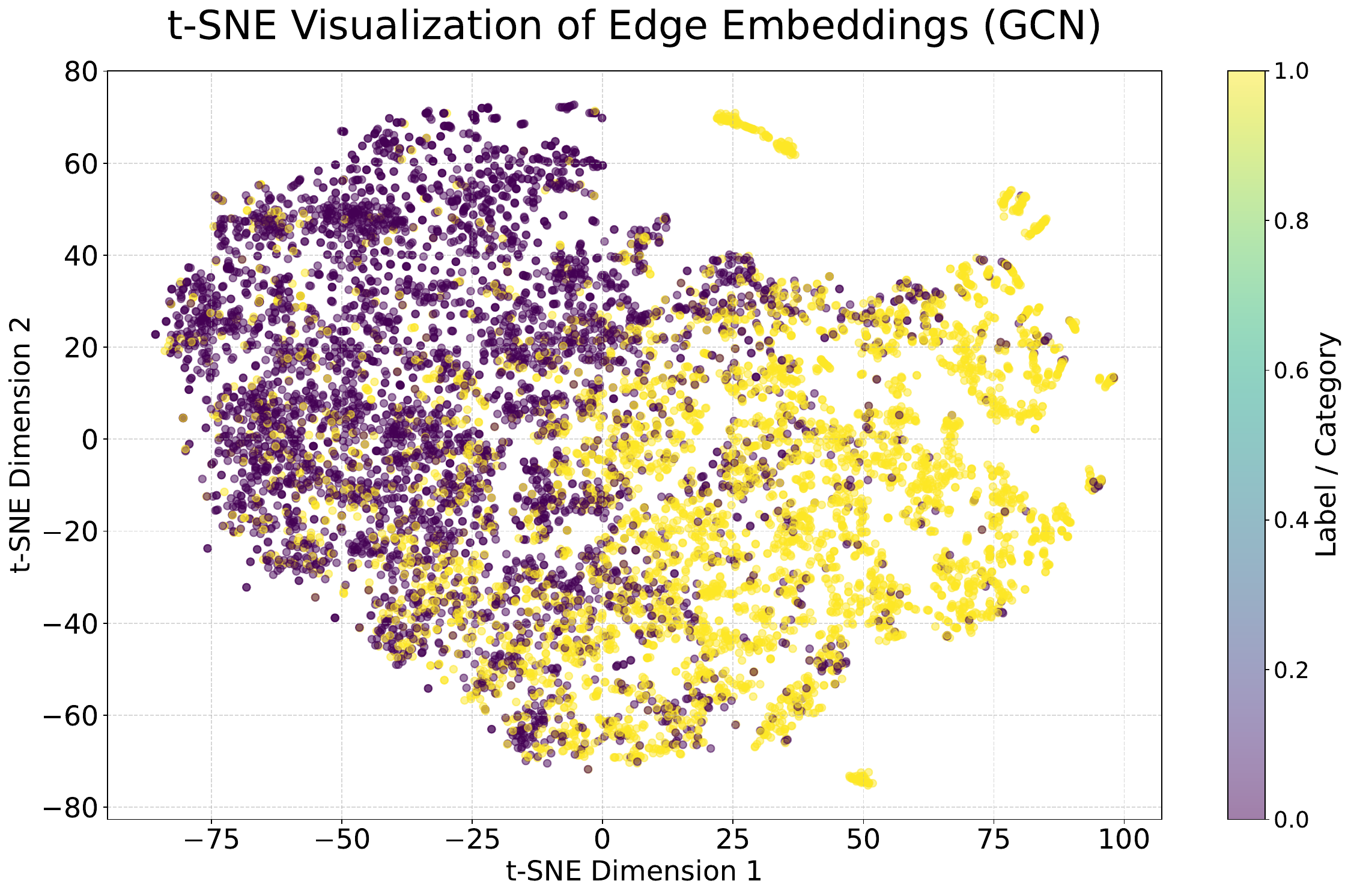}
        \caption{GCN PPT-Ohmnet}
        \label{fig:bild2_t}
    \end{subfigure}
    \hfill
    \begin{subfigure}[b]{0.23\textwidth}
        \centering
        \includegraphics[width=\textwidth]{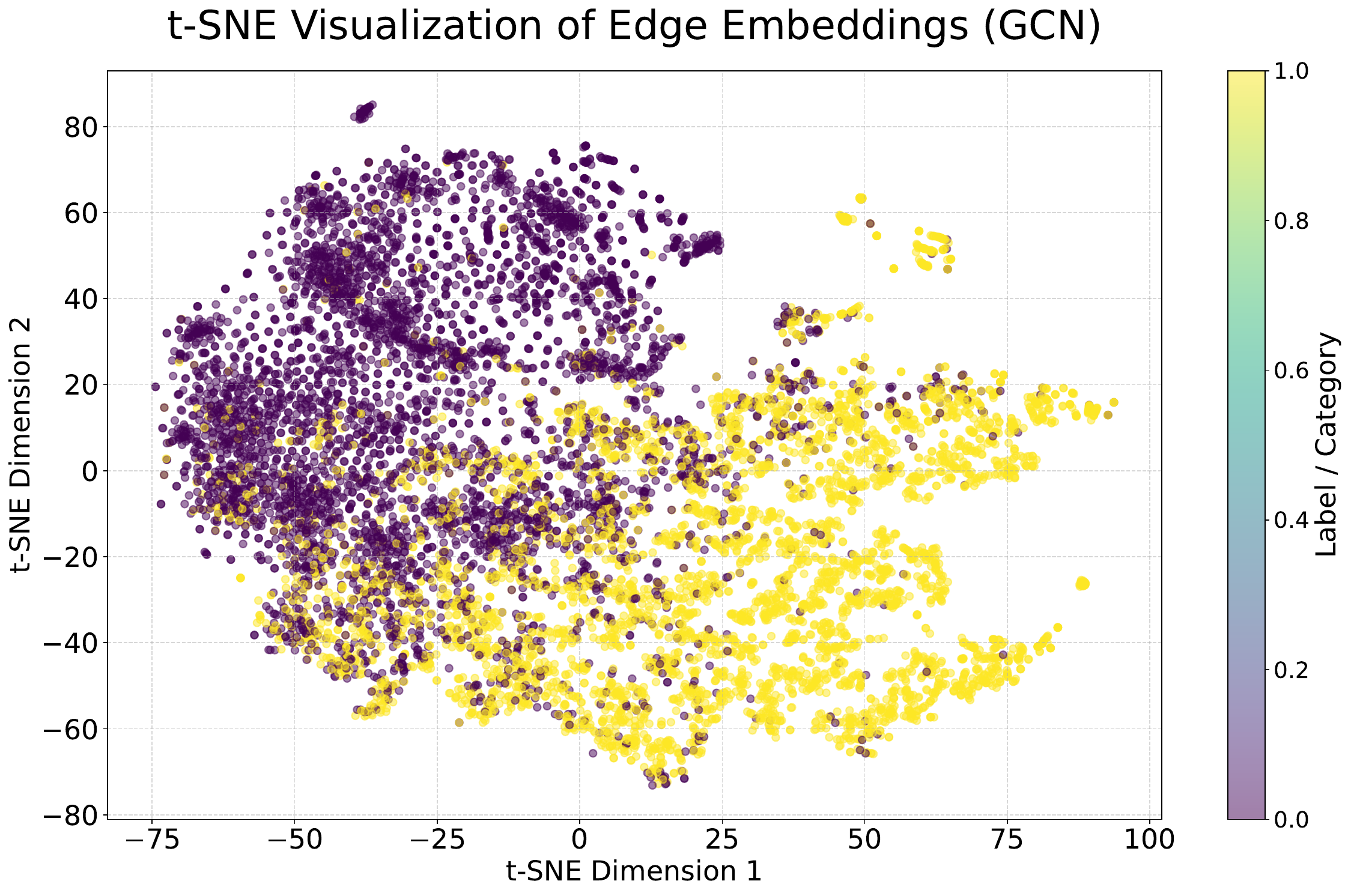}
        \caption{GCN PP-Pathways}
        \label{fig:bild3_t}
    \end{subfigure}
    \hfill
    \begin{subfigure}[b]{0.23\textwidth}
        \centering
        \includegraphics[width=\textwidth]{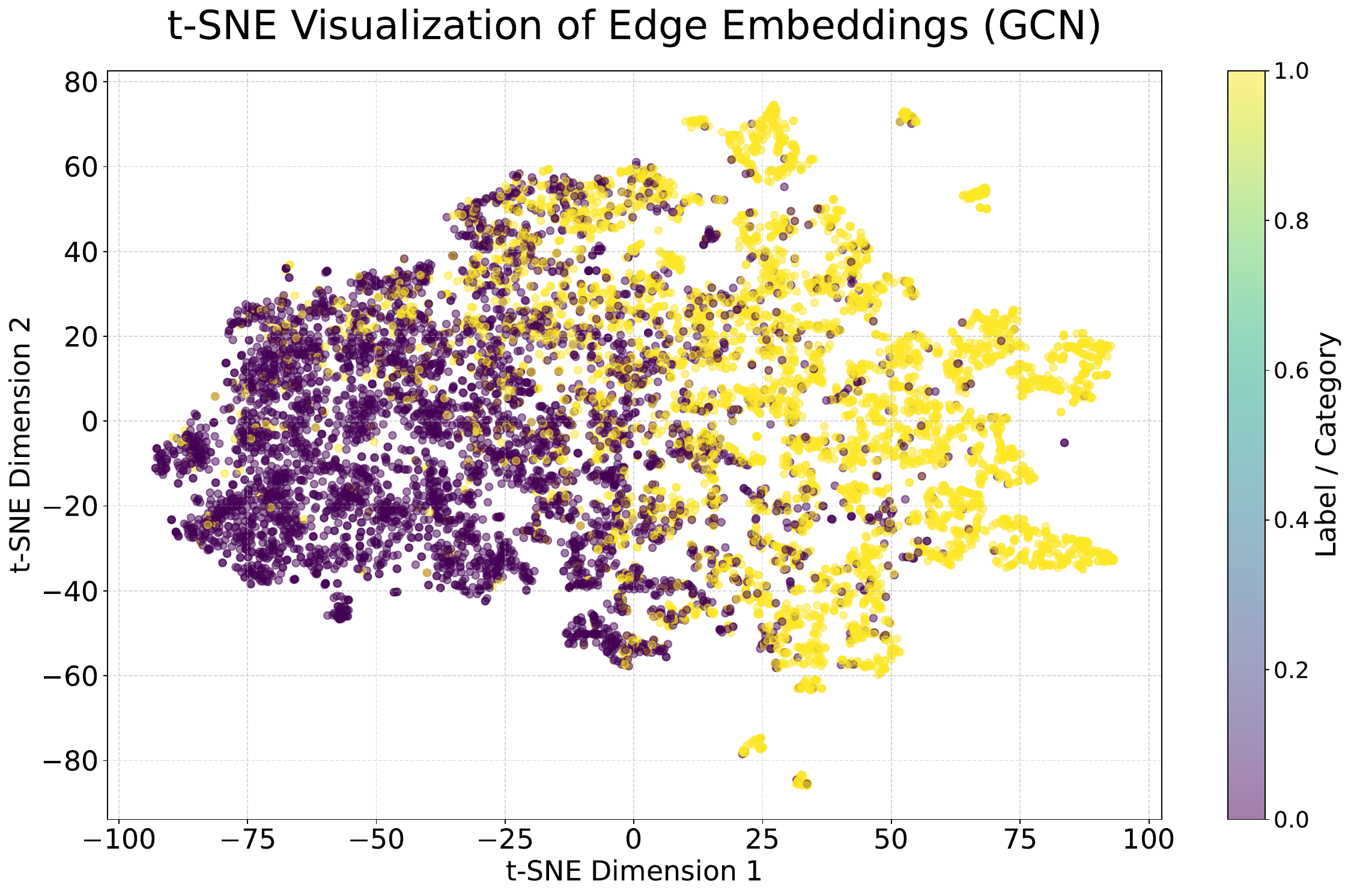}
        \caption{GCN PP-Decagon}
        \label{fig:bild4_t}
    \end{subfigure}

    \vspace{0.5cm} 

    \begin{subfigure}[b]{0.23\textwidth}
        \centering
        \includegraphics[width=\textwidth]{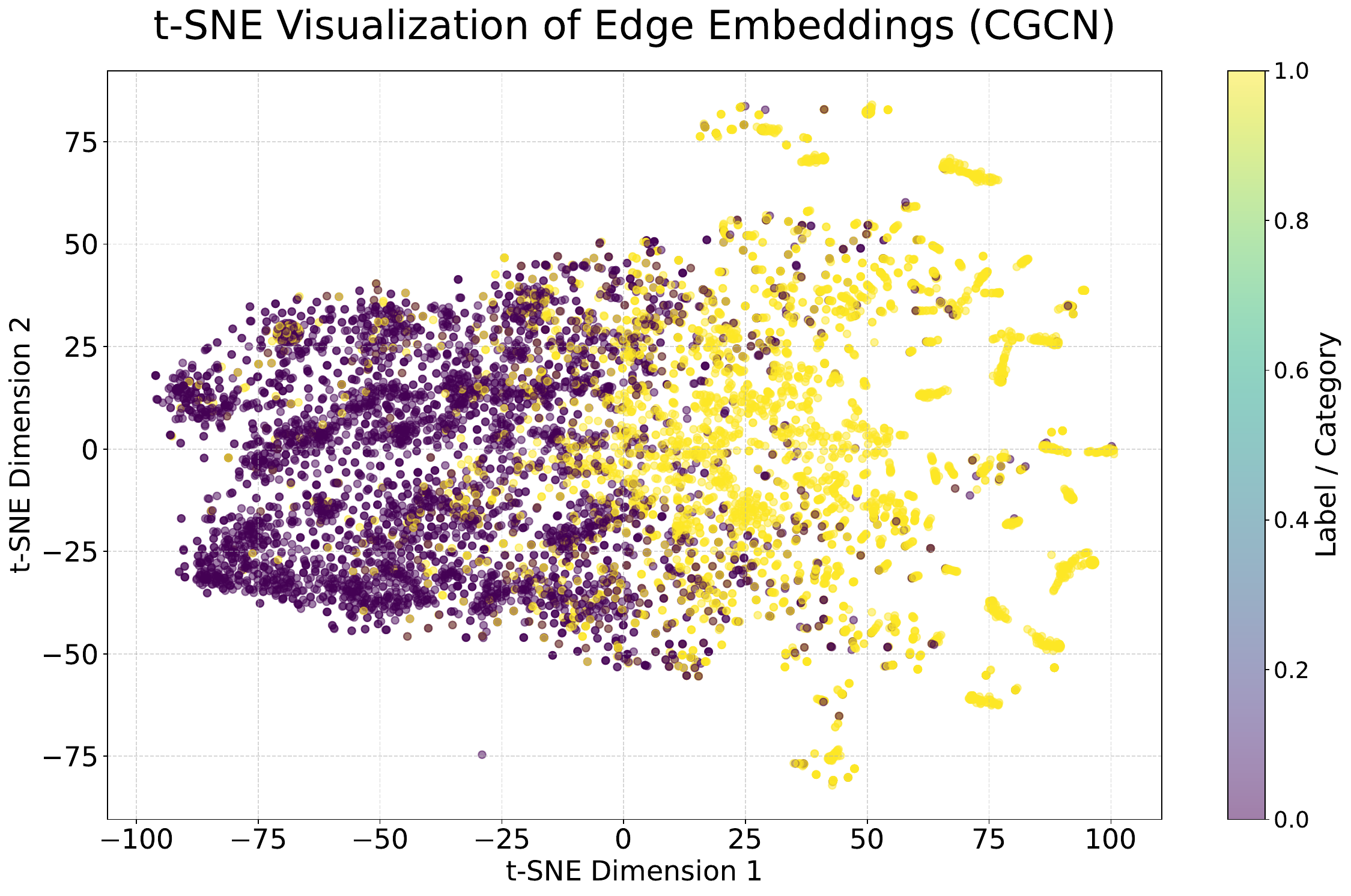}
        \caption{CGCN HuRI-PPI}
        \label{fig:bild5_t}
    \end{subfigure}
    \hfill
    \begin{subfigure}[b]{0.23\textwidth}
        \centering
        \includegraphics[width=\textwidth]{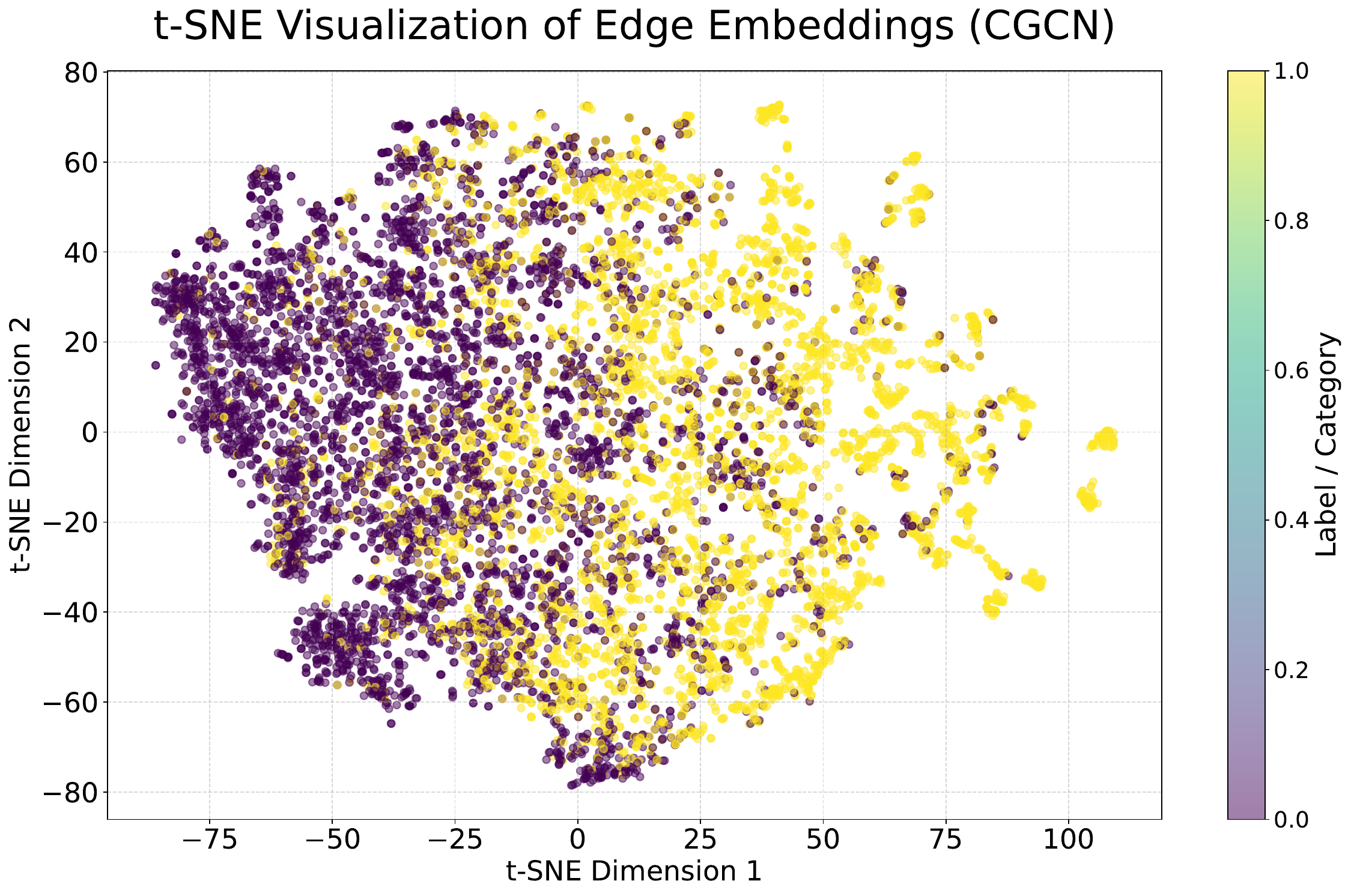}
        \caption{CGCN PPT-Ohmnet}
        \label{fig:bild6_t}
    \end{subfigure}
    \hfill
    \begin{subfigure}[b]{0.23\textwidth}
        \centering
        \includegraphics[width=\textwidth]{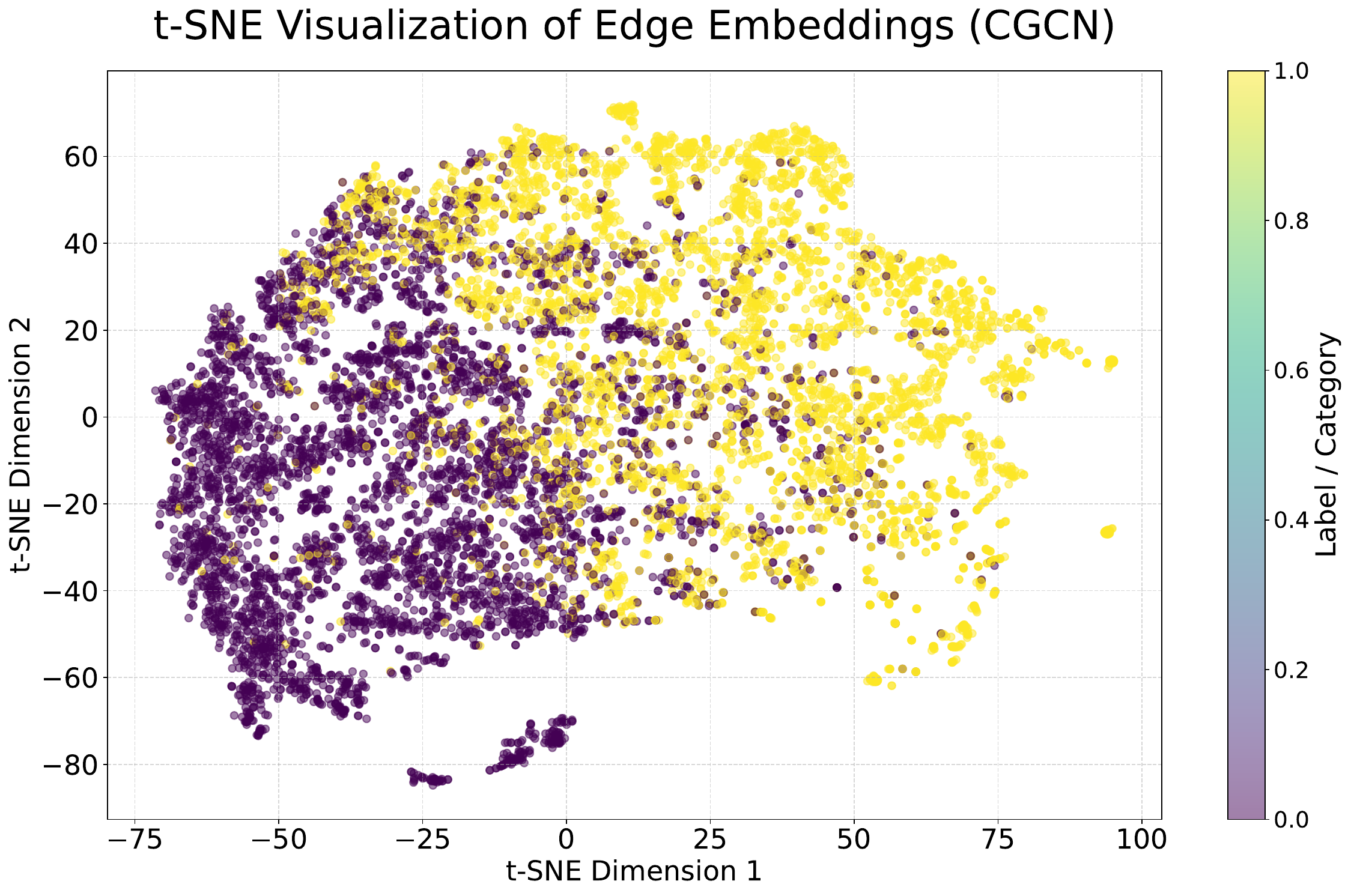}
        \caption{CGCN PP-Pathways}
        \label{fig:bild7_t}
    \end{subfigure}
    \hfill
    \begin{subfigure}[b]{0.23\textwidth}
        \centering
        \includegraphics[width=\textwidth]{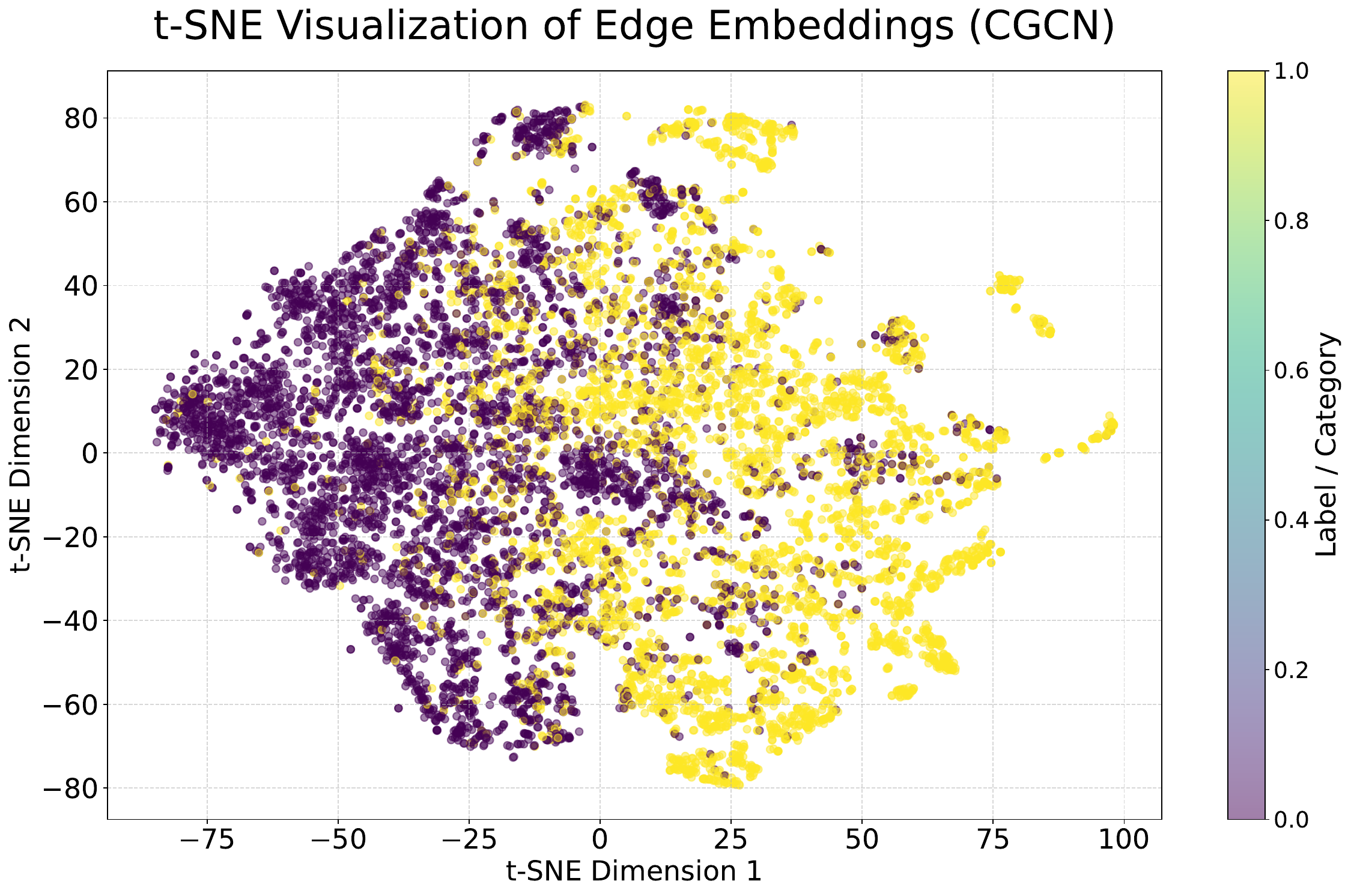}
        \caption{CGCN PP-Decagon}
        \label{fig:bild8_t}
    \end{subfigure}

    \caption{The representation vectors of each edge
in the test datasets (sample of 10000 edges for large datasets) are projected into 2D spaces by t-SNE. The yellow and purple points represent node pairs without link relationship and node pairs with link relationship, respectively.}
    \label{fig:tsne}
\end{figure}

Shen et al. \cite{shen2024curvature} study CGCN in representation learning and evaluate edge embeddings using positive and negative edges as target classes, visualized via t-SNE projections into a 2D space. Their comparison of CGCN with other network embedding models, such as DeepWalk, LINE, Node2Vec, and SDNE, shows a better separation between these two classes in CGCN. However, this analysis fails to include the representation of GCN, which does not differ strongly from the representation of CGCN. Therefore, the improved separation between positive and negative edges does not originate from the incorporation of ORC. Here, we provide this comparison of the representations of GCN and CGCN (see Figure~\ref{fig:tsne}). For a comparison with other network embedding models, refer to \cite{shen2024curvature}.

\clearpage

\end{document}